%% file: elsarticle-template-num.tex
\documentclass[preprint,12pt]{elsarticle}

\usepackage{lipsum} 
\usepackage{verbatim}

\usepackage{amssymb}
\usepackage{amsmath}
\usepackage{xcolor}
\usepackage{dsfont}
\usepackage{subcaption}
\usepackage{algorithm}
\usepackage{algpseudocode}
\usepackage{multirow}
\usepackage{hhline}

\usepackage{soul}
\definecolor{lightyellow}{rgb}{.90,.95,1}
\definecolor{lightyellow}{rgb}{1,1,0.2}
\sethlcolor{lightyellow}

\usepackage{colortbl}

\usepackage{booktabs}
\usepackage{colortbl}
\usepackage{hhline}

\journal{Medical Image Analysis}
\begin{document}

\begin{frontmatter}

\title{Exploring the robustness of TractOracle methods in RL-based tractography}
\tnotetext[]{Code is available at https://github.com/scil-vital/tractoracle-irt}

\author[1]{Jeremi Levesque}
\author[1]{Antoine Théberge}

\author[1]{Maxime Descoteaux}
\author[1]{Pierre-Marc Jodoin}

\affiliation[1]{organization={Department of Computer Science, Faculty of Science, University of Sherbrooke},
            addressline={2500 Bd de l'Université}, 
            city={Sherbrooke},
            postcode={J1N 3C6}, 
            state={Québec},
            country={Canada}}

\begin{abstract}
Tractography algorithms leverage diffusion MRI to reconstruct the fibrous architecture of the brain's white matter. Among machine learning approaches, reinforcement learning (RL) has emerged as a promising framework for tractography, outperforming traditional methods in several key aspects. TractOracle-RL, a recent RL-based approach, reduces false positives by incorporating anatomical priors into the training process via a reward-based mechanism.

In this paper, we investigate four extensions of the original TractOracle-RL framework by integrating recent advances in RL, and we evaluate their performance across five diverse diffusion MRI datasets. Results demonstrate that combining an oracle with the RL framework consistently leads to robust and reliable tractography, regardless of the specific method or dataset used.

We also introduce a novel RL training scheme called \textit{Iterative Reward Training (IRT)}, inspired by the Reinforcement Learning from Human Feedback (RLHF) paradigm. Instead of relying on human input, IRT leverages bundle filtering methods to iteratively refine the oracle’s guidance throughout training. Experimental results show that RL methods trained with oracle feedback significantly outperform widely used tractography techniques in terms of accuracy and anatomical validity.

\end{abstract}


\end{frontmatter}

\section{Introduction}

Tractography is the computational process of reconstructing fibrous tissues  using diffusion magnetic resonance imaging (dMRI) \cite{basser2000vivo}. It is primarily employed to model white matter (WM) tracts in the human brain and remains the only non-invasive and reliable method for doing so. Accurate reconstruction of major neural pathways has proven valuable for various applications, including neurosurgical planning \cite{essayed2017white, soni2017diffusion}, tractometry \cite{Takemura24}, and connectomics \cite{Henderson20, Zhang22}.

Most traditional tractography algorithms operate by iteratively propagating streamlines following local diffusion information at or near the current tracking point. These methods determine the next tracking direction either deterministically—by following the maxima of a local diffusion model—or probabilistically—by sampling from the model interpreted as an orientation probability distribution~\cite{Sarwar19}.

A whole-brain tractogram that faithfully reflects anatomical reality is critical for downstream applications. However, despite algorithmic improvements, all traditional methods tend to generate a substantial proportion of anatomically invalid streamlines, known as false positives~\cite{maier2017challenge}. This stems from the inherent difficulty of the problem: attempting to reconstruct the brain's global WM structure using only local diffusion information is fundamentally ill-posed \cite{maier2017challenge}. In complex regions, such as fiber crossings or fanning areas~\cite{Jeurissen13}, local cues are often insufficient to guide the tracking in a way that ensures both local coherence and global anatomical plausibility.

To tackle this issue, global tractography algorithms have been proposed~\cite{mangin2013toward}. These methods tend to minimize a cost function based on the alignment of streamline segments in each voxels~\cite{christiaens2015global,kreher2008gibbs}. While appealing in theory, these methods tend to suffer from prohibitive computational costs. Moreover, the difficulty of imposing anatomical priors means that even in a \emph{in silico} context without noise, these methods may not surpass traditional tractography~\cite{maier2017challenge}.

Given that maintaining local coherence and global anatomical plausibility is a challenging task in tractography, post-processing techniques have been developed to filter out false positive streamlines. Examples include SIFT \cite{Smith13, Smith15}, COMMIT \cite{Daducci15}, extractor\_flow \cite{Petit23}, Verifyber \cite{astolfi2023supervised} and FINTA \cite{LEGARRETA2021}. A common strategy is to over-seed the brain—generating a surplus of streamlines—in the hope that, after filtering, most regions will retain enough anatomically plausible connections. However, this approach has notable drawbacks. First, filtering methods themselves are not infallible and may be prone on filtering out rare, but valid, tracts, and second, over-seeding can significantly increase computational demands and runtime. Furthermore, in regions that are inherently difficult to track and yield a low proportion of true positives, filtering alone cannot recover missing streamlines and improve anatomical accuracy.

Supervised ML approaches have shown promise in addressing some limitations of traditional tractography methods \cite{neher2017fiber, Poulin17, Cai24, Poulin19, wasserthal2018tractseg}. These techniques require models to be trained on annotated streamlines, which involves manual validation by domain experts such as neuroanatomists and radiologists. However, this process is highly impractical for large-scale datasets due to the significant time and expertise required. Even assuming unlimited resources, the inherent variability between expert annotations makes it difficult to establish a reliable ground truth~\cite{Rheault20}. As a result, most datasets suitable for supervised learning in tractography, which are based on ground truths, are limited to in silico phantoms such as the FiberCup~\cite{Cote13} and the ISMRM-2015 dataset~\cite{maier2017challenge}, which may not fully reflect the complexity of human anatomy.
Some \textit{in vivo} datasets (e.g. TractoInferno~\cite{poulin2022tractoinferno}, HCP105~\cite{wasserthal2018tractseg}) were built to provide a standard structure and also provide a reference that can be used to train machine learning algorithms. In the case of TractoInferno, the reference streamlines were obtained using silver standard tractography algorithms as well as silver standard filtering algorithms which all have their own biases towards an accurate reconstruction of the anatomy. HCP105 has been semi-automatically annotated (as it involved some manual quality control and cleanup), but still suffer from incomplete regions (i.e. partially or fully missing bundles) due to using silver standard tractography algorithms. The inherent limitations of the reference streamlines essentially form a performance ceiling (w.r.t the anatomy) for downstream ML algorithms trained on such datasets.

Reinforcement learning (RL) has emerged as a viable alternative to supervised methods for tractography~\cite{theberge2021track, theberge2024matters, theberge2024tractoracle}. Unlike supervised learning, RL does not require reference streamlines which circumvents the limitations of the reference tracts from the in vivo datasets described above. Instead, a reward function guides the learning process of a so-called tractography agent. Early RL-based approaches used reward functions to mimic traditional algorithms—for instance, by rewarding alignment with the local dMRI signal orientation~\cite{theberge2021track}. However, this formulation inherits the same limitations of local-only methods i.e., the inability to capture global anatomical context.

Moreover, stepwise reward structures are prone to reward hacking~\cite{Skalse22}. Since the agent receives rewards at each tracking step, it may learn to generate unnecessarily long streamlines that continue following plausible local directions—maximizing cumulative reward without ensuring anatomical validity. This behaviour frequently led to a significant number of false positives, despite the agent successfully optimizing its training objective.

To overcome these limitations, Théberge et al.~\cite{theberge2024tractoracle} introduced {\em TractOracle}, a novel framework that integrates global anatomical knowledge into the RL training process using a so-called {\em oracle} neural network. This oracle is a transformer-based model trained on a silver standard set of streamlines, and is designed to evaluate the anatomical plausibility of candidate streamlines.

Once trained, the oracle is embedded into the RL framework in two key ways. First, it contributes to the reward function by providing a positive reward whenever the agent generates anatomically valid streamlines, encouraging more meaningful exploration during training. Second, during inference, the oracle acts as a stopping criterion—monitoring each in-progress streamline and terminating tracking if the growing path becomes implausible. This dual role allows the oracle to serve as both a guide and a gatekeeper, helping reduce the prevalence of anatomically invalid streamlines.

Empirical results demonstrated that this strategy substantially lowers the false positive rate, validating the effectiveness of incorporating non-local anatomical priors into the RL training pipeline.

This paper expands upon and reinforces the preliminary findings presented in TractOracle \cite{theberge2024tractoracle}, offering a deeper analysis and broader empirical validation. We show that the core framework—integrating anatomical knowledge into the training process via a neural network oracle—is highly robust regardless of the RL-based tractography settings or any specific implementation choices. Whether updating the RL algorithm to a more modern variant, altering the number of streamline points passed to the oracle, enriching the agent’s input with broader local context or jointly training the oracle and the agent, the framework consistently outperforms state-of-the-art baselines in all scenarios. Additionally, with minor refinements to the agent's training procedure, we further improve tracking accuracy, solidifying the framework’s lead over comparable methods.

\begin{samepage}
The main contributions of this paper are the followings:
\begin{itemize}
    \item We thoroughly validate the TractOracle framework with a variety of RL algorithms;
    \item We tackle the challenges of local tractography by endowing agents with a broader context window;
    \item We propose a new training scheme called {\em Iterative Reward Training} for RL-based tractography, improving the anatomical validity of reconstructed tractograms in vivo.
\end{itemize}
\end{samepage}

\section{RL framework for tractography}
\subsection{Tractography}

Tractography is the process of reconstructing WM pathways from dMRI. In this work we explore \emph{iterative} tractography, where pathways (henceforth called \emph{streamlines}, or \emph{tracts}), represented by an ordered set $P = \{p_0 \ldots p_T\}$ where $ p_t\in\mathbb{R}^3$, are obtained iteratively. From an initial 3D point $p_0$, the diffusion model (here a fiber ODF (fODF)~\cite{Tournier04, Descoteaux09}), a vector field $v$ is evaluated and a direction $\mathbf{a}_0 \sim v(p_0)$ of amplitude $\Delta$ is selected. The process iterates as $p_{t+1} = p_t + \Delta \mathbf{a}_t$ until a stopping criterion is met, such as reaching the maximum streamline length, encountering a sharp angle between $\mathbf{a}_{T-1}$ and $\mathbf{a}_T$, or stepping outside the white matter.

\subsection{Reinforcement learning}\label{sec:rl}
RL models are commonly formalized as Markov Decision Processes (MDPs), where an agent $\pi$ interacts with its environment through a set of states $S$, actions $A$, transition probabilities $p(\mathbf{s}_{t+1} | \mathbf{s}_t, \mathbf{a}_t)$, and rewards $r(\mathbf{s}_t, \mathbf{a}_t)$, abbreviated as $r_t$.  The reward $r_t$ can be seen as a return value that is high when the action $a_t$ taken at state $s_t$ is good and low when it is not.

In the context of tractography, the environment corresponds to the 3D diffusion signal—specifically, fODFs. The agent $\pi_\theta$ is modelled as a neural network parameterized by $\theta$, which takes as input the state $\mathbf{s}_t$, a vector encoding the local environment around the current position $p_t$, and predicts an action $\mathbf{a}_t$, representing a direction of motion. The predicted action $\mathbf{a}_t \sim \pi_\theta(\mathbf{s}_t)$ leads to a new state $\mathbf{s}_{t+1}$ and a reward $r_t$. 

By iteratively applying this process, a trajectory is formed, and the objective becomes to optimize the policy $\pi_\theta$ to maximize the expected discounted return:
\begin{equation}
G_t = \sum_{k=t}^{T} \gamma^k r(\mathbf{s}_{t+k}, \mathbf{a}_{t+k}),
\end{equation}
where $\gamma \in [0,1]$ is the discount factor, which favors immediate rewards when $\gamma \rightarrow 0$. The value function $V_{\pi_\theta}(\mathbf{s}_t)$ estimates the expected return from state $\mathbf{s}_t$, while the action-value function (Q-function) $Q_{\pi_\theta}(\mathbf{s}_t, \mathbf{a}_t)$ estimates the expected return of taking action $\mathbf{a}_t$ in state $\mathbf{s}_t$ before following policy $\pi_\theta$:
\begin{align}
V_{\pi_\theta}(\mathbf{s}_t) &= \mathbb{E}_{\pi_\theta} [G_t \mid \mathbf{s}_t], \\
Q_{\pi_\theta}(\mathbf{s}_t, \mathbf{a}_t) &= \mathbb{E}_{\pi_\theta} [r_t + G_{t+1} \mid \mathbf{s}_t, \mathbf{a}_t].
\end{align}

Optimizing $\pi_\theta$ constitutes the core of RL research, with a wide range of algorithms proposed in the literature. One particularly effective algorithm for tractography applications is Soft Actor-Critic (SAC)~\cite{haarnoja2018sac}, which was shown to be well suited for this domain~\cite{theberge2024matters}. SAC augments the standard RL objective by incorporating the policy's entropy, promoting exploration and avoiding premature convergence:
\begin{equation}
G_t = \sum_{k=t}^{T} \gamma^k \left[ r(\mathbf{s}_{t+k}, \mathbf{a}_{t+k}) + \alpha \mathcal{H}\left( \pi_\theta(\cdot \mid \mathbf{s}_{t+k}) \right) \right],
\end{equation}
where $\alpha$ is a learned temperature parameter controlling the entropy term's influence. The agent collects transitions in a replay buffer $B$ and uses it to train both its Q-functions (parametrized by two neural networks, namely $\phi_1 \text{ and } \phi_2$) and its policy $\pi_{theta}$ (also a neural network parametrized by $\theta$) at each step. The target networks $\phi_1'$ and $\phi_2'$ are used to compute the Bellman residual.

See \ref{sec:sac-appendix} for a detailed explanation of Soft Actor-Critic. The reward will be described in the next subsections.

\subsection{TractOracle}

Theberge et al.~\cite{theberge2024tractoracle} introduced a reward function with two components: a local term promoting the alignment between $a_t$, the fODF maxima $\overline{v(p_t)}$ at $p_{t}$ and $a_{t-1}$, and an anatomical score of the reconstructed streamline.  Formally: 
\begin{align}
    r_t \, = \, 
    \underbrace{
    \big(|\max_{\overline{v(p_t)}}{\langle \overline{v(p_t)}, {a}_{t} \rangle| \cdot \langle a_t, a_{t-1}\rangle}\big)}_{local} + 
    \alpha 
    \underbrace{\mathds{1}_{\Omega_\psi}(P_{0..t})}_{anatomical} \\
     \mathds{1}_{\Omega_\psi}(P_{0..t}) = \begin{cases}
        1 & \text{if \;} \Omega_\psi(P_{0..t}) \geq 0.5 \text{\; and \;} t = T \\
        0 & \text{else,}
    \end{cases}   
\end{align}
with $\Omega_\psi$ \emph{TractOracle-Net}, a transformer-based neural network~\cite{vaswani2017attention} trained to assess the anatomical validity of streamlines. In other words, when a P-long streamline $P_{0..t}$ is deemed anatomically plausible, i.e. when TractOracle-Net $\Omega_\psi(P_{0..t}) \geq 0.5 $, the reward at step $t$ is augmented by a factor $\alpha$.

TractOracle-Net, which we will refer to as \textit{an oracle}, is a transformer with 4 transformer encoding blocks each having 4 multi-attention heads, totalling 550K parameters. For a detailed architecture illustration, please refer to \ref{sec:oracle-architecture}.

\section{Methodology}

In order to demonstrate how effective the use of an oracle is in RL-based tractography, we introduce several improvements to {\em TractOracle} to align it with recent best practices in RL. Specifically, we generalize the framework to support a broader range of RL algorithms, enhance the state representation, and implemented a strategy to better align the reward function with the learned policies.  These methods are described in the following four subsections.  The reader may also find further technical details in the supplementary material from \ref{app:rl-details} to \ref{sec:fodf-ae-impl}.

\subsection{Oracle}\label{sec:oracle-workflow}

In this work, we expand upon the experimental procedure of~\cite{theberge2024tractoracle} by training oracles and agents on mutiple datasets and using multiple filtering methods as reference. Namely, on the in silico dataset (c.f. section~\ref{sec:datasets}), we use local and PFT~\cite{girard2014towards} tracking to generate reference streamlines which are then resampled to a fixed number of points and then annotated by the Tractometer~\cite{renauld2023validate} as true or false positives. On the in vivo HCP dataset~(c.f.~\ref{sec:datasets}), we again use local and PFT~\cite{girard2014towards} tracking but instead use Verifyber, extractor\_flow and RecobundlesX to provide three annotations per streamline, which are then used to train three oracles. Finally, on the in vivo TractoInferno dataset, we repeat the tracking procedure of~\cite{theberge2024tractoracle} and annotate the streamlines using the previous three methods. In all cases, to train the oracles, we split the annotated reference streamlines in training (80\%), validation (10\%) and testing (10\%) sets. In addition to filtering methods, we also explore how varying the number of points used to resample streamlines affects the prediction accuracy. Indeed, Théberge et al~\cite{theberge2024tractoracle} trained their oracle using streamlines resampled to 128 points. However, previous work has shown that streamlines can be effectively compressed to far fewer points—as few as 12—without significant loss of information~\cite{garyfallidis2012quickbundles,rheault2020analyse}. Given that transformer models relying on self-attention scale quadratically with sequence length~\cite{vaswani2017attention}, and that TractOracle-Net is invoked at every tracking step, using shorter sequences is computationally advantageous. In this work, we therefore consider oracles trained with streamlines resampled to 32 points. We additionally explore the impact of the number of points per streamlines in section~\ref{sec:exp-oracle-nb-points}.  Oracles were trained using a mean squared error loss. For an brief list of the hyperparameters used, refer to table~\ref{tab:hyperparams-oraclenet}.

\subsection{DroQ}

While SAC has been the defacto model-free RL algorithm since its proposal, several papers~\cite{janner2019trust, chen2021randomized, hiraoka2021dropout} have outlined its sample inefficiency: many transitions need to be sampled from the environment to properly train agents. Numerous improvements over SAC have been proposed over the years. One such method is Dropout Q-Functions (DroQ)~\cite{hiraoka2021dropout}, a reinforcement learning algorithm that builds on SAC. Essentially, DroQ performs training updates more frequently while collecting fewer transitions which improves SAC's sample efficiency. As naively increasing the number of updates can destabilize learning, the authors introduce additional tricks to address this issue. Additional details on the modifications done are detailed in \ref{sec:droq-appendix}. Experimental results demonstrate that these modifications lead to higher returns with fewer environment interactions compared to SAC and other related methods.

\subsection{CrossQ}

CrossQ~\cite{bhatt24} is another approach aimed at improving the SAC algorithm. Developed concurrently with DroQ, CrossQ seeks to simplify SAC while enhancing its sample efficiency. To achieve this, the authors remove the target critic networks and carefully incorporate Batch Normalization (BatchNorm) layers~\cite{bhatt24} into the Q-networks. 

A key detail in their approach is the treatment of state-action pairs: tuples $(s_k, a_k)$ and $(s_{k+1}, a_{k+1})$ are concatenated \emph{batch-wise} to ensure they are processed as part of the same distribution. This design allows both to contribute equally to BatchNorm's running statistics, which helps stabilize training. The authors argue that these simplifications not only simplify the architecture but also lead to improved performance and sample efficiency.

For more details on CrossQ, please refer to~\ref{sec:crossq-appendix}.

\subsection{Widening neighbourhood signal}
Previous work on the formulation of the state $\mathbf{s}_t$ in RL-based tractography~\cite{theberge2024matters, sinzinger2022reinforcement} has primarily focused on \emph{local} representations—typically limited to a small neighborhood of voxels surrounding the current position $p_t$. However, relying solely on local information exacerbates the inherently ill-posed nature of tractography. To mitigate this limitation, we explore expanding the spatial context provided to the agent by increasing the radius of voxels included in the input.

\begin{figure}[t]
    \centering
        \includegraphics[width=0.7\textwidth]{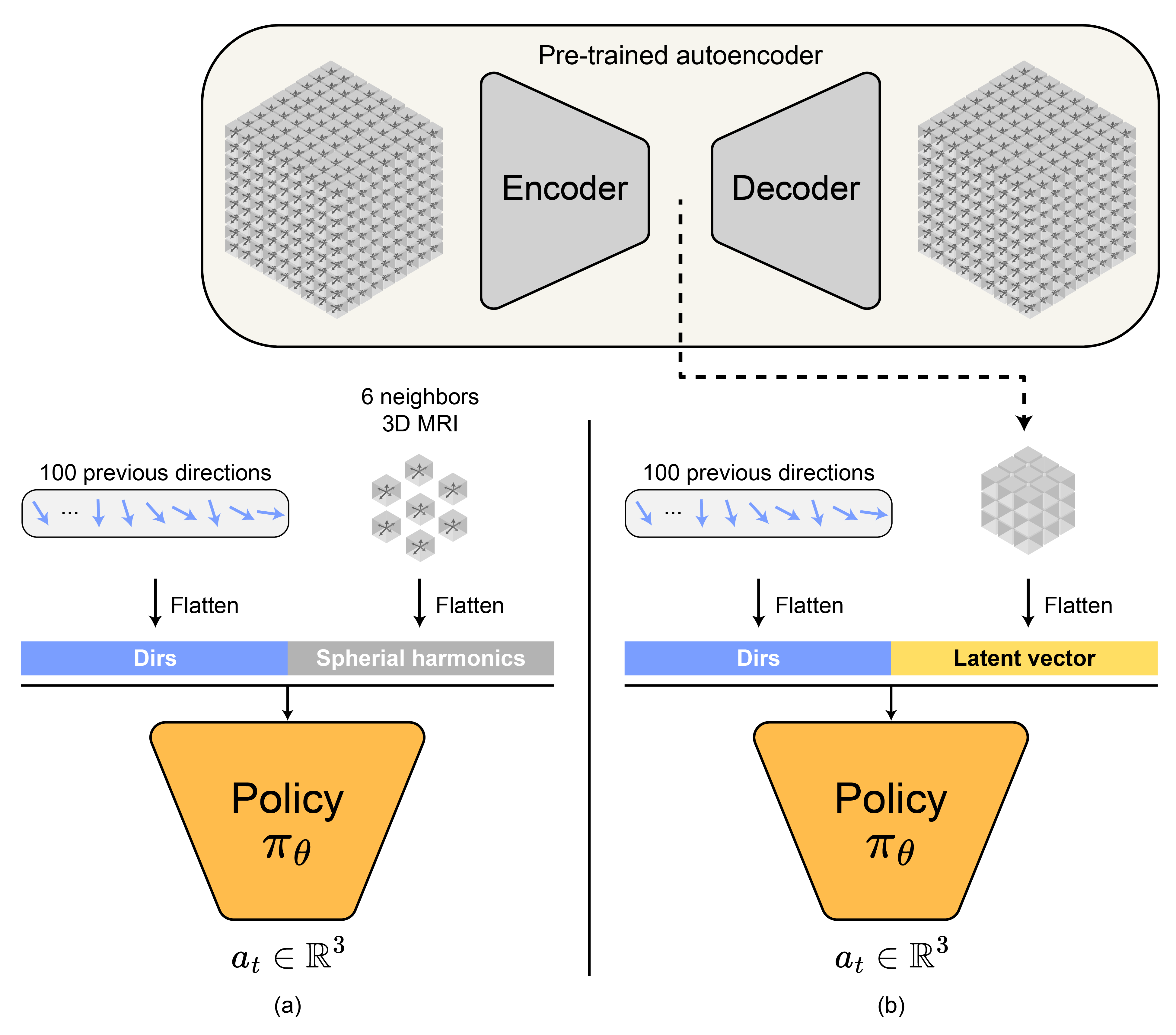}
    \vspace{-0.2cm}
    \caption{Different state formulations. (a) State formulation as presented in~\cite{theberge2021track} using the spherical harmonics to the 6 closest neighbours of $p_t$. (b) An encoder encodes a volume of $N\times N\times N$ voxels each having 28 diffusion MRI spherical harmonics coefficients into a $3\times 3\times 3\times 32$ tensor that is flattened and given to the agent as a vector of length 864. In both cases, the 100 last tracking steps are included~\cite{theberge2024matters}. Details about the training of the autoencoder as well the encoder and decoder architectures are provided in \ref{sec:fodf-ae-impl}}.
    \label{fig:fodf-ae}
\end{figure}

Earlier methods, including {\em TractOracle}~\cite{theberge2024tractoracle}, use only the six immediate neighbors of $p_t$—corresponding to the adjacent voxels in the up, down, left, right, anterior, and posterior directions (see figure\ref{fig:fodf-ae})(a)). Given that each voxel encodes 28 spherical harmonics coefficients, this results in a feature vector of size $28 \times 7 = 196$.

In this work, we incorporate a larger 4D window of size $N \times N \times N \times 28$ centered at $p_t$. As illustrated in figure~\ref{fig:fodf-ae} (b), this voxel cube is processed by an convolutional encoder that outputs a 4D tensor of shape $3 \times 3 \times 3 \times 32$, where 32 denotes the number of output channels. This tensor is then flattened into a 1D vector of length $864$, which is used as input to the agent. The sheer reason for using an encoder is to reduce the size of the input data as we empirically found that a raw $N\times N\times N\times 28$ bloc leads to poor results when $N>5$.

Consistent with the {\em TractOracle} architecture, both state formulations concatenate this vector with the previous 100 tracking directions to form the final state $\mathbf{s}_t$.

\subsection{Iterative reward training (IRT)}

It is well known that different tracking algorithms produce  different looking streamlines, and consequently streamlines from different algorithms may have a different \emph{shape distribution}. Empirically, we found that streamlines produced by RL-based algorithms follow this trend, including TractOracle-RL.

We therefore hypothesize that TractOracle-Net, the streamline evaluation network introduced in TractOracle~\cite{theberge2024tractoracle} trained solely on streamlines tracked by classical methods, may suffer from~\emph{distributional shift} when scoring streamlines produced by TractOracle-RL.

\begin{figure}
    \centering
    \includegraphics[width=0.9\linewidth]{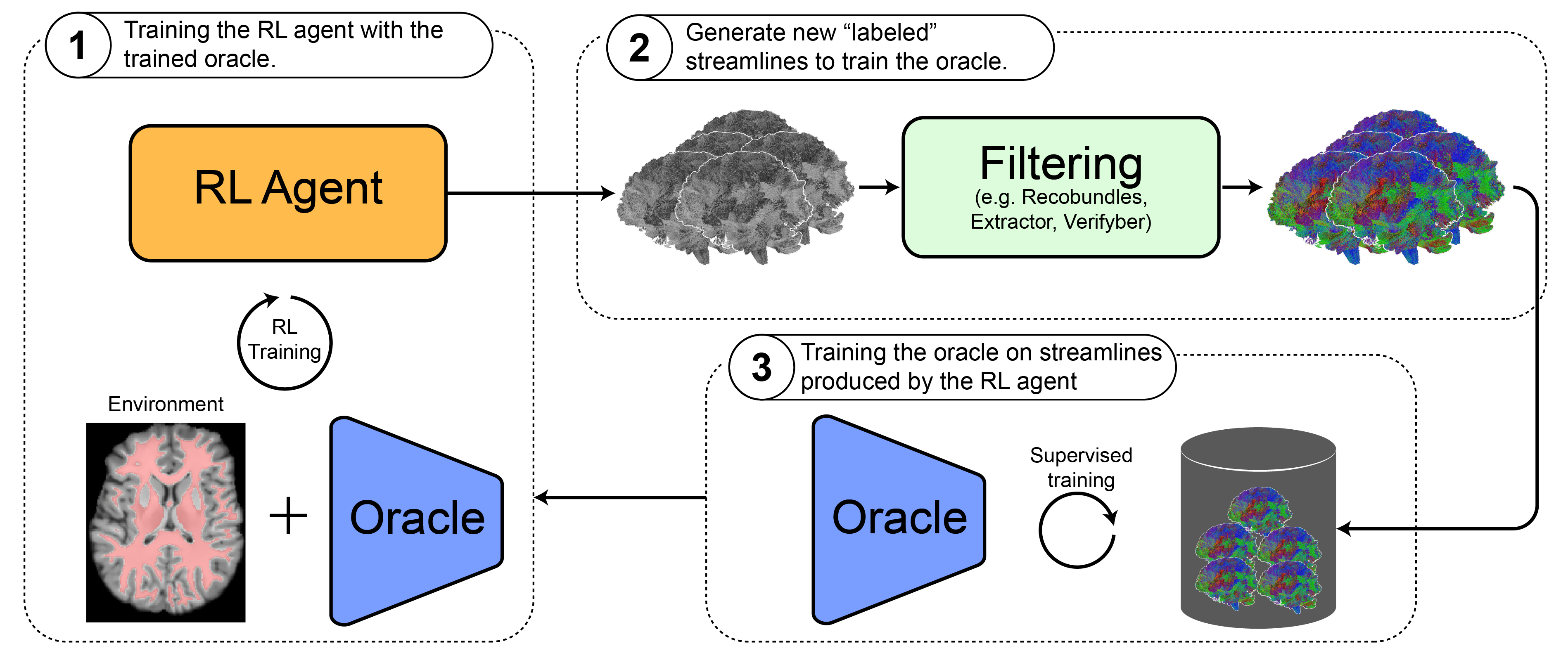}
    \caption{The 3-step Iterative Reward Training (IRT) method. \textbf{Step 1.} The RL agent is trained for a limited number of epochs. \textbf{Step 2.} The agent generates whole-brain tractograms for a set of training subjects randomly selected from the dataset. The resulting streamlines are classified as being anatomically plausible or implausible using a filtering method. The streamlines generated at each iteration are accumulated in a growing dataset. \textbf{Step 3.} The oracle is fine-tuned for a few epochs using the updated dataset, after which the process returns to Step~1.}
    \label{fig:ira-process}
\end{figure}

\begin{figure}[htbp]
    \centering
    \begin{subfigure}[t]{0.48\linewidth}
        \centering
        \includegraphics[height=3.40cm]{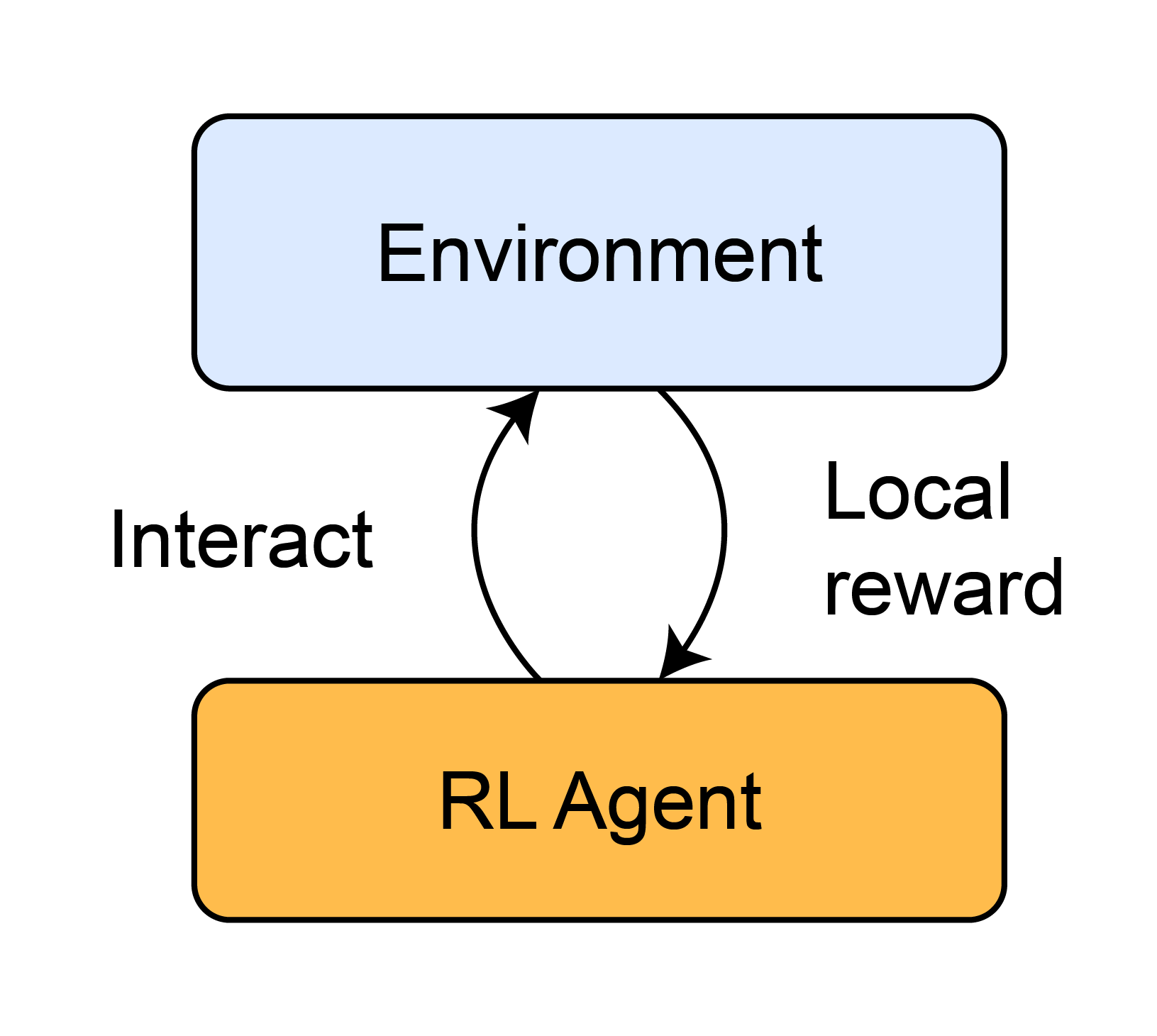}
        \caption{}
        \label{figsum:tracktolearn}
    \end{subfigure}
    \begin{subfigure}[t]{0.48\linewidth}
        \centering
        \includegraphics[height=3.40cm]{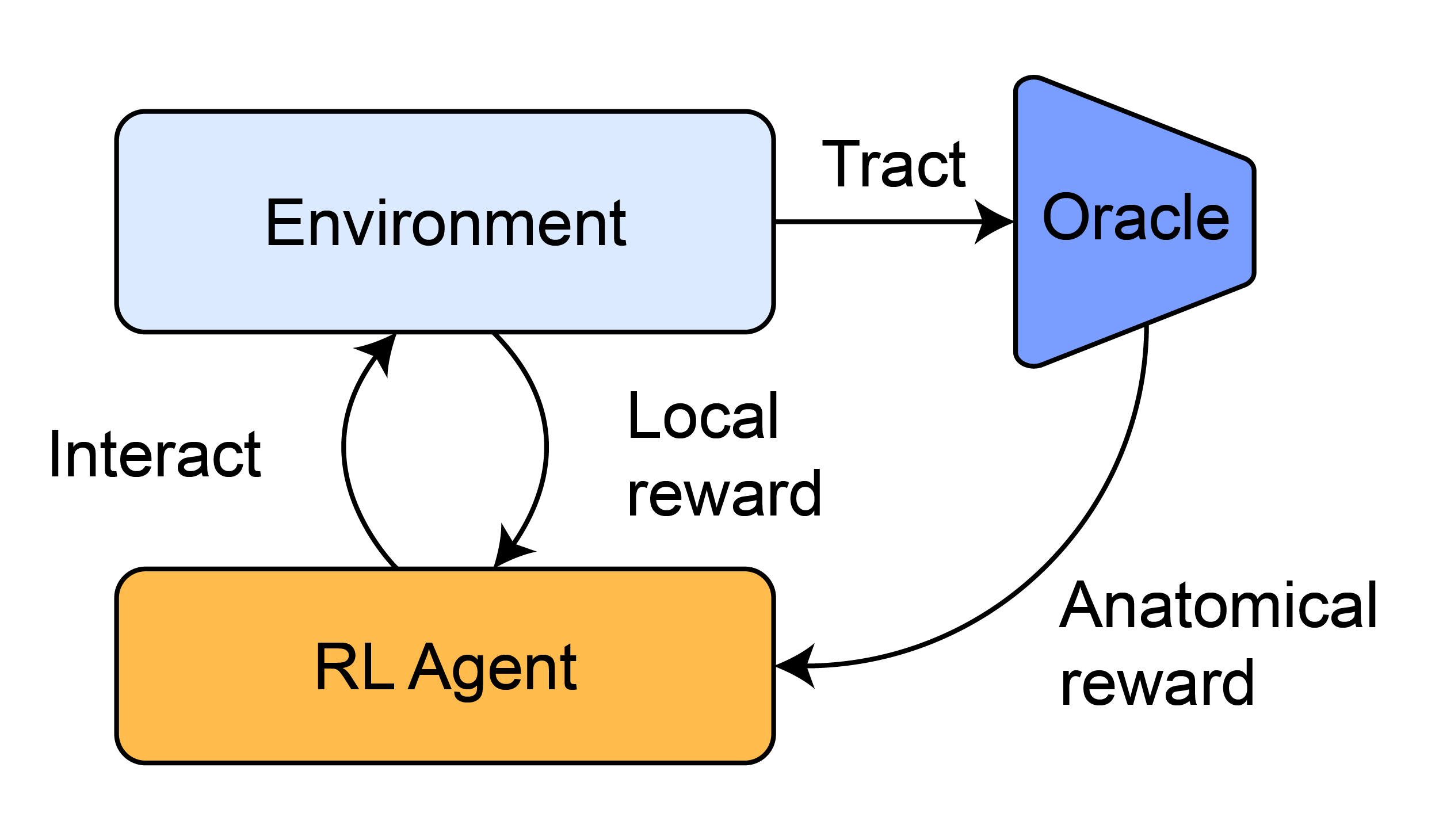}
        \caption{}
        \label{figsum:tractoracle}
    \end{subfigure}

    \begin{subfigure}[t]{1.\linewidth}
        \centering
        \includegraphics[height=3.40cm]{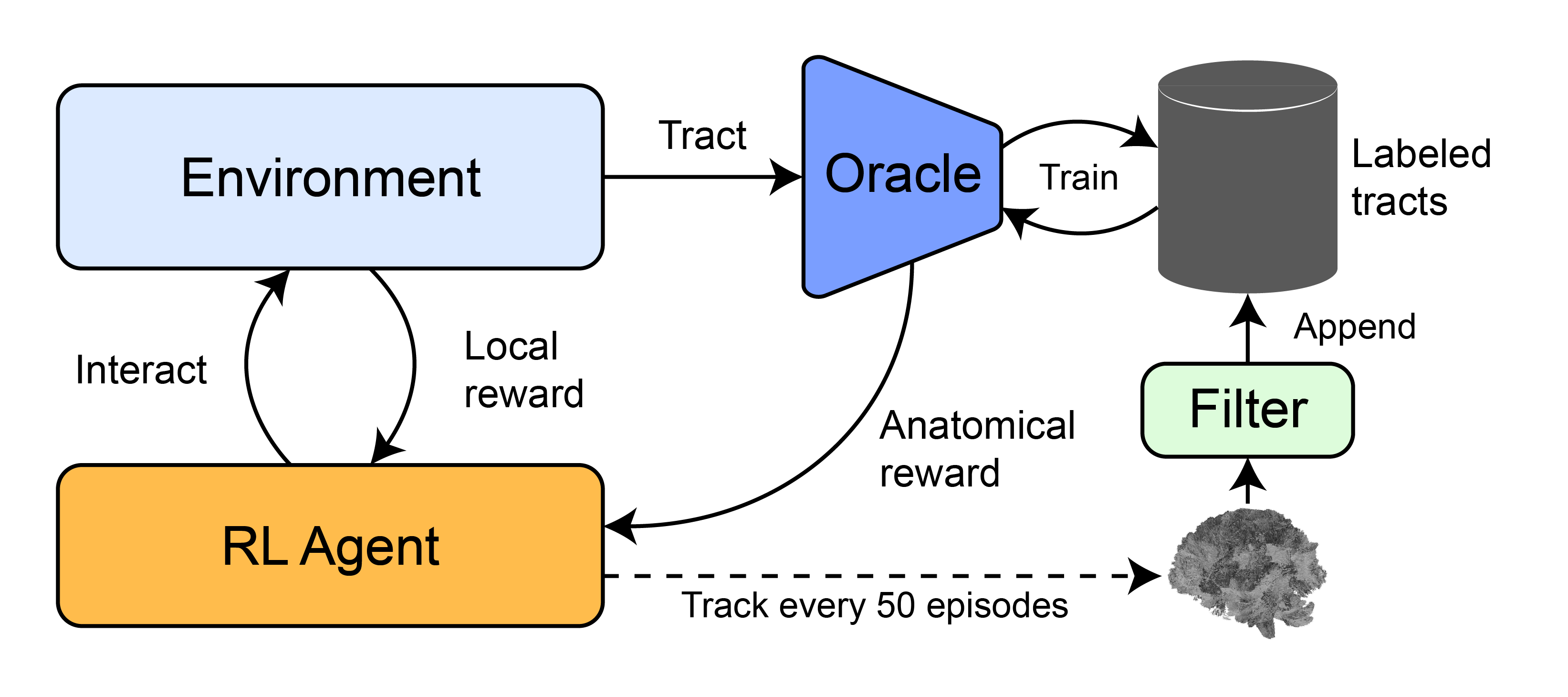}
        \caption{}
        \label{figsum:irt}
    \end{subfigure}
    \caption{Evolution of RL-based tractography training paradigms. \textbf{(a)} Track-to-Learn (TTL): the agent trains by optimizing the local alignment reward at each tracking step. \textbf{(b)} TractOracle: the agent is train to jointly optimize the local reward and the additional anatomically-informed reward. \textbf{(c)} Iterative reward tuning: the reward model is iteratively aligned with the filtering method.}
    \label{figsum:rl-methods-evolution}
\end{figure}

In order to tackle this distribution shift, we propose to iteratively train the oracle jointly with the RL agent, inspired by Reinforcement Learning from Human Feedback~(RLHF) methods (like ChatGPT)~\cite{RAY23}. We name this process \textit{iterative reward training} (IRT) and refer the reader to figure~\ref{fig:ira-process} for a step-by-step illustration of IRT and to figure~\ref{figsum:rl-methods-evolution} for a comparison with other RL-based tractography methods. This procedure aims to \textit{align} the distribution of streamlines produced by RL-tracking methods with the distribution of streamlines used to trained the oracle. We aim to keep the performance of the oracle so that it is on-par or better with what was measured during its initial training process.

IRT is a three-step method. First, the RL agent $\pi_\theta$ is trained with the initial oracle until it reaches a plateau (about 150-200 episodes). Then, the agent generates tractograms for a subset of randomly-picked subjects (here 5 subjects) from the training dataset until approximatively 250K streamlines are accumulated using the current policy. The tractograms are then filtered by RecobundlesX~\cite{garyfallidis2018recognition}, extractor\_flow~\cite{Petit23} or Verifyber~\cite{astolfi2023supervised} so their streamlines are annotated as plausible or implausible, similar to the workflow described in section \ref{sec:oracle-workflow}. The annotated tracts are then split into training, validation and testing sets each holding 80\%, 10\%, 10\% of the examples before being appended to the dataset which includes annotated tracts from previous iterations.. Finally, the oracle is trained for a few epochs. Then, the RL training loop goes back to step 1 where the tracking agent is trained some more. The process is repeated until the agent has performed 3000 episodes of training.

On the first IRT iteration, we train the oracle for 5 epochs so it can quickly recover a good prediction accuracy on tractograms generated by RL-based tractography algorithms. On subsequent IRT iterations, we train the oracle for a single epoch as we train the oracle relatively often compared to the rate of behaviour change of the RL agent. In other words, the shape distribution of the streamlines should change only slightly in between each IRT iteration and we hypothesize that training for a single epoch is done to essentially maintain the good prediction accuracy during the whole RL agent training.

We limit the size of the dataset to 4M streamlines, leading to the "earlier" streamlines being overwritten after a few iterations. As the training improves the performance of the agent, the agent should be more and more biased towards reconstructing plausible (i.e. valid) streamlines. With this in mind, before appending new data to the dataset, we randomly sample the over represented class to match the number of streamlines of the under represented class, which keeps the dataset balanced with 50\% plausible and 50\% implausible streamlines. The data was carefully split in training, validation and testing sets in order to keep each separate dataset balanced so that we're never in a scenario with any class that is over-represented.

\subsection{Tracking agents}\label{sec:tracking_methods}
The policy $\pi_\theta$ as well as the Q-networks and the value functions are all four-layer multilayer perceptrons with three hidden linear layers of size 1024 and a final output layer of size 3 or 1 for the policy and the critic respectively.

To assess the impact of extended training, we also evaluated a variant trained for 3,000 episodes, denoted \emph{SAC-3K}, to observe whether prolonged training could potentially cause divergence or reward hacking problems. Additional variants include \emph{DroQ-1K} and \emph{CrossQ-3K}, which respectively incorporate the DroQ and CrossQ RL algorithms. \emph{CrossQ-3K} is trained for 3,000 episodes while we trained DroQ-1K for 1,000 episodes due to its slower run time and higher sample efficiency. We also introduce \emph{CrossQ-AE}, a CrossQ-based variant that leverages the latent space encoding described in Figure~\ref{fig:fodf-ae}(b).

Finally, we implement two versions of TractOracle trained with the iterative reward training (IRT) scheme: \emph{SAC-IRT} and \emph{CrossQ-IRT}.

In order to quantify and effectively evaluate each tracking experiment, we use different strategies to assess the performance of the different methods. Upon training and tracking on in silico datasets, we utilize the Tractometer~\cite{renauld2023validate} to provide 8 different metrics (which are listed and detailed in \ref{sec:tractometer-metrics}) that we use to compare the different algorithms. However, since the in vivo datasets do not provide ground truth tract annotations, we leverage RecobundlesX~\cite{garyfallidis2018recognition,rheault2020analyse}, extractor\_flow~\cite{Petit23} and Verifyber~\cite{astolfi2023supervised} to quantify the number of streamlines recognized by each method.

Since \textit{in vivo} datasets do not provide ground truth streamlines or tract annotations, for all methods
\section{Experiments and results}

\subsection{Datasets}\label{sec:datasets}

Five datasets were used in the following experiments.

\begin{enumerate}[1)]
    \item \textbf{ISMRM2015}
    is synthetic dataset generated from a single subject from the Human Connectome Project ~\cite{maier2017challenge,Maier-Hein15}. Global tractography was performed on the subject and 25 reference bundles were manually segmented. FiberFox~\cite{neher2014fiberfox} was used to generate a synthetic diffusion volume (2mm iso, 32 directions, b=1000) and T1w image from the segmented bundles. We repeat the experimental procedure of Theberge et al.~\cite{theberge2024tractoracle} and tracked $\sim$1M streamlines with five classical tractography algorithms~\cite{tournier2019mrtrix3, girard2014towards, st2018surface} using each's default set of parameters, seeding in both the WM and the WM-GM interface. Streamlines were then input to the Tractometer~\cite{renauld2023validate} to obtain true positives and false positives, acting as labelled data to train the oracles.

    \item \textbf{BIL \& GIN}~\cite{mazoyer2016bil} is a publicly-available dataset of 453 healthy adults mostly in their twenties.  All acquisitions were performed on a Philips Achieva 3T scanner using 21 non-colinear diffusion gradient directions, a b-value of 1,000 s/mm², four signal averages, and an isotropic resolution of 2 mm. Following the experimental setup by Legaretta et al.~\cite{LEGARRETA2021}, we focus on the callosal streamlines of 39 subjects; specifically on homotopic streamline bundles—those connecting gyral-based segment pairs across hemispheres—within 26 predefined anatomical regions, which are treated as plausible anatomical connections for training TractOracle-Net, while everything else is treated as unplausible.  For more details on these bundles and how they were extracted, please refer to section A5 of Legaretta et al.~\cite{LEGARRETA2021}.

    \item \textbf{TractoInferno}~\cite{poulin2022tractoinferno} is a publicly accessible and highly diverse dataset developed to support machine learning applications in diffusion MRI (dMRI) tractography. It includes data from 354 individuals, aggregated from six distinct sources and acquired using five different MRI scanners, encompassing a wide range of resolutions, acquisition protocols, and participant age groups. All MRI scans underwent manual quality control prior to tractography, which was performed using an ensemble of four methods: local deterministic, local probabilistic, PFT~\cite{girard2014towards}, and SET~\cite{st2018surface}. The resulting streamlines were then processed using RecoBundlesX to produce a silver-standard reference.

    \item \textbf{Human Connectome Project (HCP)} is a collection of \textit{in vivo} subjects, of which we randomly selected 100. Each subject's diffusion image was acquired at 1.25mm iso using 288 directions over three shells (b=1000, 2000, 3000)~\cite{Sotiropoulos13}. Each subject was processed using TractoFlow~\cite{theaud2020tractoflow} to obtain fODFs, peaks, tissue segmentations and T1w images registered in the diffusion space. We use 80 subjects for training, 10 for validation and 10 for testing.

    \item \textbf{Penthera 3T}~\cite{paquette2019penthera} is a dataset of 12 healthy individuals of an average age of $25.92 \pm 1.86$. Each subject was scanned 6 times: 3 times per session for 2 sessions. Each diffusion image was acquired at 2.0mm iso over three shells (b=300, 1000, 2000) on 8, 32, and 60 distributed directions respectively. Each subject was using TractoFlow~\cite{theaud2020tractoflow} to obtain fODFs, peaks, tissue segmentations and T1w images registered in the diffusion space at a resolution of 1mm iso. From this dataset, we use the first two scans of the first session for the 12 subjects for a total of 24 acquisitions.
\end{enumerate}

For all datasets, we use fODFs of order 6 (28 spherical harmonics coefficients) as fODFs of order 8 were too computationally expensive and order 4 were too low frequency. FODFs of order 6 ended up being a good compromise between a good representation and computational efficiency. Furthermore, we used 6 to replicate the experimental setup of previous RL tractography method that we compare our methods to~\cite{theberge2021track,theberge2024tractoracle}.

\subsection{Experiment 1: number of points}\label{sec:exp-oracle-nb-points}

For the first experiment, we explore the impact of the number of points per streamline used to train oracles. We consider streamlines resampled to 32, 64 and 128 points on the ISMRM2015 and BIL\&GIN datasets.

\subsection{Experiment 2: in silico tracking}

In this experiment, we evaluate the performance of our RL tracking agents, SAC, DroQ, CrossQ and their variants on the ISMRM2015 dataset using the metrics extracted by the Tractometer. We perform an experiment on a synthetic dataset first to essentially conclude whether our methods and their variants are effective candidates for \textit{in vivo} tracking. It is easier to quickly compare and analyze the generation performance of different algorithms using a well-defined ground truth that provides extensive metrics.

We compare our methods against three well-known non-ML tractography algorithms, namely \textbf{sd\_stream} and \textbf{iFOD2} from MRTrix3~\cite{tournier2019mrtrix3} and Parallel Transport Tractography~\cite{aydogan2020parallel} (\textbf{PTT}) from Trekker as well as two reinforcement learning-based approaches: the original \textbf{Track-to-Learn} (TTL) method~\cite{theberge2021track} and the more recent \textbf{TractOracle} framework~\cite{theberge2024tractoracle}, referred to as SAC-1k from now on.

\subsection{Experiment 3: in vivo tracking}

In this experiment, we compare the same tracking methods as the previous experiment, but on in vivo images where we can truly assess the generation capabilities of our algorithms on real-world data. The objective of this experiment is to evaluate which methods yield a higher number of true positive streamlines. All machine learning-based tracking methods were trained and tested on both the TractoInferno and HCP datasets using 20 seeds per voxel. From this point onwards, we omit DroQ and CrossQ-AE as their training is significantly slower and the tracking speed of CrossQ-AE is prohibitively slow as we highlight in section \ref{sec:ismrm-results} below.

\subsection{Experiment 4: transfer learning}

In this experiment, we assess the generalization capability of our agents by testing them on entirely different datasets without performing any additional fine-tuning. To ensure consistency with the prior evaluations, we generate the tractograms using the same procedure as in the previous section-specifically, using 20 seeds per voxel.

All experiments are run using a single GPU, either a NVidia RTX4090 or a NVidia V100SXM2, as the training of most models does not require more than 4Gb of VRAM, with the exception of training the oracle where 32Gb of VRAM is recommended. Although training our experiments require the use of GPUs, utilizing the trained models afterwards is much less memory-consuming and can easily be performed on a consumer laptop.

\subsection{Results}

\subsubsection{Scoring streamlines resampled to fewer number of points}\label{sec:oracle-nb-points}

Results on the ISMRM2015 and BIL\&GIN datasets are presented in Table~\ref{t:oracle-ismrm2015}. Reducing the number of points per streamline does not degrade oracle classification performance, but significantly reduces inference time. Based on these findings—and since no performance drop was observed when using TractOracle-Net-32 instead of TractOracle-Net-128 during TractOracle-RL training—we adopted the 32-point streamlines.

\begin{table}[tp]
    \centering
    \footnotesize
    \caption{Classification metrics on the ISMRM2015 and BIL\&GIN datasets. RecobundlesX and FINTA results are reported from \cite{legarreta2021filtering}.
    }
    \begin{tabular}{lccccc}
        \hline
        \multicolumn{6}{c}{\textbf{ISMRM2015}}\\
        \hline
        & \textbf{Accuracy} & \textbf{Sensitivity} & \textbf{Precision} & \textbf{F1-score} & \textbf{Time{\scriptsize (s)}} \\
        \hline
        RecobundlesX     & 0.91 & 0.81 & 0.97 & 0.88 & - \\
        FINTA           & 0.91 & 0.91 & 0.91 & 0.91 & -\\
        \hline
        \hline
        TractOracle-Net-128 & \textbf{0.98} & 0.98 & 0.97 & \textbf{0.98} & 21 \\
        TractOracle-Net-64 & \textbf{0.98} & \textbf{0.99} & 0.97 & \textbf{0.98} & 10 \\
        TractOracle-Net-32 & \textbf{0.98} & 0.98 & \textbf{0.98} & \textbf{0.98} & \textbf{6} \\
        \hline
    \end{tabular}
    \begin{tabular}{lccccc}
        \hline
        \multicolumn{6}{c}{\textbf{ BIL\&GIN }}\\
        \hline
        & \textbf{Accuracy} & \textbf{Sensitivity} & \textbf{Precision} & \textbf{F1-score} & \textbf{Time{\scriptsize (s)}} \\
        \hline
        RecobundlesX     & 0.82 & 0.80  & 0.67 & 0.70  & - \\
        FINTA           & 0.91 & \textbf{0.91} & \textbf{0.78} & \textbf{0.83}  & - \\
        \hline
        \hline
        TractOracle-Net-128 & 0.95 & \textbf{0.91} & 0.74  & 0.82  & 20 \\
        TractOracle-Net-64  & 0.95  & \textbf{0.91} & 0.75 & 0.82   & 10 \\
        TractOracle-Net-32  & \textbf{0.96} & \textbf{0.91} & 0.76 & \textbf{0.83} & \textbf{6} \\
        \hline
    \end{tabular}
    \label{t:oracle-ismrm2015}
\end{table}

\subsubsection{In silico results : ISMRM2015}\label{sec:ismrm-results}

The Track-to-Learn variants introduced in section~\ref{sec:tracking_methods} were first evaluated on the in silico dataset. Results are presented in Table~\ref{tab:ismrm} -please refer to the supplementary material for a detailed description of the eight evaluation metrics. 

\begin{table}[tp]
    \centering
    \footnotesize
    \caption{Tractometer scores on ISMRM2015. Scores in \textbf{bold} indicate the best method for each metric, in {\color{blue}blue} are methods that severely underperformed and \underline{underlined} is the best method with an oracle at their core. 1K and 3K stand for the number of training episodes. For conciseness, all methods below the double horizontal line (i.e. SAC, CrossQ, DroQ) refer to a TractOracle experiment. \ref{sec:tractometer-metrics} offers a detailed list of the acronyms used to identify the metrics in this table.}
    \hfill
    \begin{tabular}{lcccc}
        \hline
        & \textbf{VC \% $\uparrow$} & \textbf{VB (/21) $\uparrow$} & \textbf{IC \% $\downarrow$} & \textbf{IB $\downarrow$} \\
        \hline
        sd\_stream & 55.96 $\pm$ 0.21 & 19.00 $\pm$ 0.00 & {\color{blue}44.04 $\pm$ 0.21} & 199.80 $\pm$ 4.26 \\
        ifod2 & {\color{blue}31.53 $\pm$ 0.20} & 19.00 $\pm$ 0.00 & {\color{blue}68.47 $\pm$ 0.20} & {\color{blue}281.00 $\pm$ 4.00} \\
        ptt & {\color{blue}26.41 $\pm$ 0.05} & 19.40 $\pm$ 0.55 & {\color{blue}54.44 $\pm$ 0.03} & {\color{blue}457.40 $\pm$ 4.62} \\
        TTL~\cite{theberge2021track} & 66.13 $\pm$ 1.15 &\textbf{20.00 $\pm$ 0.63} & {\color{blue}33.87 $\pm$ 1.15} & {\color{blue}293.40 $\pm$ 11.8} \\
        \hline
        \hline
        SAC-1K~\cite{theberge2024tractoracle} & 88.05 $\pm$ 0.35 & 19.00 $\pm$ 0.47 & 11.95 $\pm$ 0.35 & 195.67 $\pm$ 4.99 \\
        SAC-3K & 90.73 $\pm$ 0.05 & 19.00 $\pm$ 0.00 & 9.08 $\pm$ 0.05 & 151.20 $\pm$ 2.05 \\
        DroQ-1K & 84.80 $\pm$ 0.05 & 19.00 $\pm$ 0.00 & 14.86 $\pm$ 0.05 & 171.20 $\pm$ 3.27 \\
        CrossQ-3K & \textbf{\underline{91.64} $\pm$ 0.04} & 19.00 $\pm$ 0.00 & \textbf{\underline{8.11} $\pm$ 0.03} & 159.40 $\pm$ 4.51 \\
        CrossQ-AE & 85.88 $\pm$ 0.03 & 17.00 $\pm$ 0.00 & 13.66 $\pm$ 0.04 & \textbf{\underline{142.00} $\pm$ 4.06} \\
        SAC-IRT & 89.17 $\pm$ 0.03 & 19.00 $\pm$ 0.00 & 10.58 $\pm$ 0.03 & 173.60 $\pm$ 3.13 \\
        CrossQ-IRT & 84.45 $\pm$ 0.06 & \underline{19.33} $\pm$ 0.00 & 15.00 $\pm$ 0.07 & 172.40 $\pm$ 7.99 \\
        \hline

        \hline
        & \textbf{OL \% $\uparrow$} & \textbf{OR $\downarrow$} & \textbf{F1 \% $\uparrow$} & \textbf{NC $\downarrow$} \\
        \hline
        sd\_stream & 38.85 $\pm$ 0.05 & \textbf{3.59 $\pm$ 0.05} & 52.47 $\pm$ 0.03 & {\color{blue}9.36 $\pm$ 0.18} \\
        ifod2 & 48.70 $\pm$ 0.11 & 8.18 $\pm$ 0.19 & 59.10 $\pm$ 0.07 & {\color{blue}12.45 $\pm$ 0.08} \\
        ptt & \textbf{75.32 $\pm$ 0.14} & 32.45 $\pm$ 1.43 & \textbf{65.75 $\pm$ 0.18} & {\color{blue}19.15 $\pm$ 0.05} \\
        TTL~\cite{theberge2021track} & 53.84 $\pm$ 2.28 & 29.94 $\pm$ 2.00 & 57.43 $\pm$ 1.84 & 2.85 $\pm$ 0.42 \\
        \hline
        \hline
        SAC-1K~\cite{theberge2024tractoracle} & \underline{48.43} $\pm$ 0.64 & 17.68 $\pm$ 1.00 & \underline{57.00} $\pm$ 0.47 & 0.73 $\pm$ 0.12 \\
        SAC-3K       & 35.43 $\pm$ 0.03 & 15.25 $\pm$ 0.29 & 45.87 $\pm$ 0.05 & \textbf{\underline{0.19} $\pm$ 0.01} \\
        DroQ-1K      & 38.24 $\pm$ 0.05 & 21.10 $\pm$ 0.15 & 47.00 $\pm$ 0.07 & 0.34 $\pm$ 0.01 \\
        CrossQ-3K    & 33.20 $\pm$ 0.08 & 15.53 $\pm$ 0.11 & 43.27 $\pm$ 0.08 & 0.25 $\pm$ 0.01 \\
        CrossQ-AE  & 25.63 $\pm$ 0.05 & 23.61 $\pm$ 0.15 & 31.04 $\pm$ 0.09 & 0.46 $\pm$ 0.01 \\
        SAC-IRT      & 35.77 $\pm$ 0.12 & \underline{15.05} $\pm$ 0.13 & 45.85 $\pm$ 0.15 & 0.25 $\pm$ 0.01 \\
        CrossQ-IRT   & 33.90 $\pm$ 0.10 & 15.26 $\pm$ 0.25 & 44.22 $\pm$ 0.11 & 0.54 $\pm$ 0.01 \\
        \hline
        \vspace{-1cm}
    \end{tabular}
    \label{tab:ismrm}
\end{table}

As shown in the table, methods that do not rely on an oracle (i.e., sd\_stream, ifod2, ptt and Track-to-Learn) achieve top performance on some metrics —namely, the number of valid bundles (VB), overlap (OL), overreach (OR), and F1-score—while performing poorly on others, such as valid and invalid connections (VC, IC), invalid bundles (IB), and the number of no connections (NC).

In contrast, the oracle-based methods exhibit more balanced performance across all metrics. They also achieve highly competitive results on key indicators, including the proportion of valid and invalid connections (VC + IC), the number of valid bundles (VB), the number of invalid bundles (IB), and, most notably, the percentage of no connections (NC).

Furthermore, it is important to highlight that the training of \textit{DroQ} and \textit{CrossQ-AE} is significantly slower—by a factor ranging from 10$\times$ to 30$\times$ compared to the other methods. Additionally, the tracking speed of \textit{CrossQ-AE} is prohibitively slow. For these reasons, and given that neither method demonstrated superior performance, we decided not to include them in the subsequent experiments.

\begin{figure}[t]
    \centering

    \includegraphics[width=0.48\linewidth]{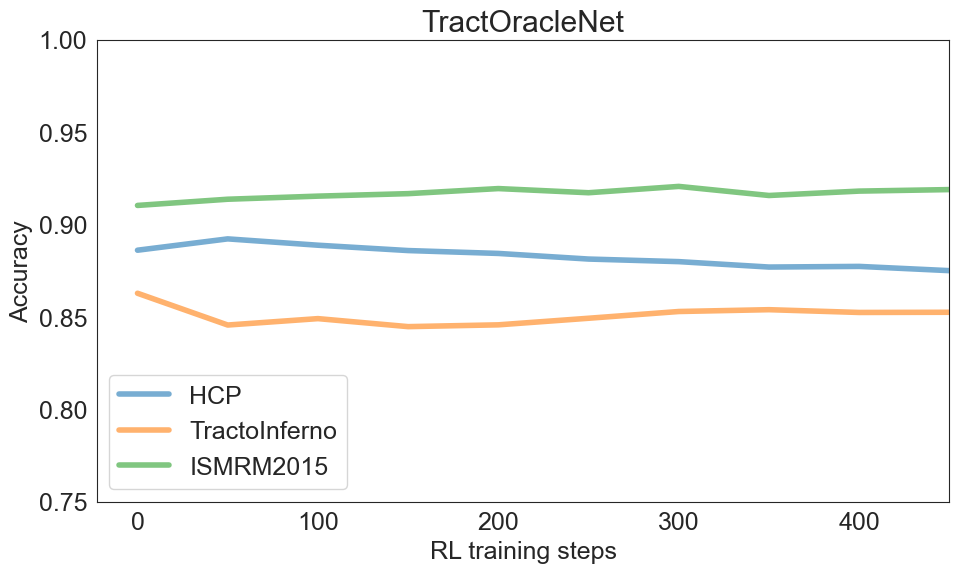}
    \label{fig:rlhf-argument}
    ~
    \includegraphics[width=0.48\linewidth]{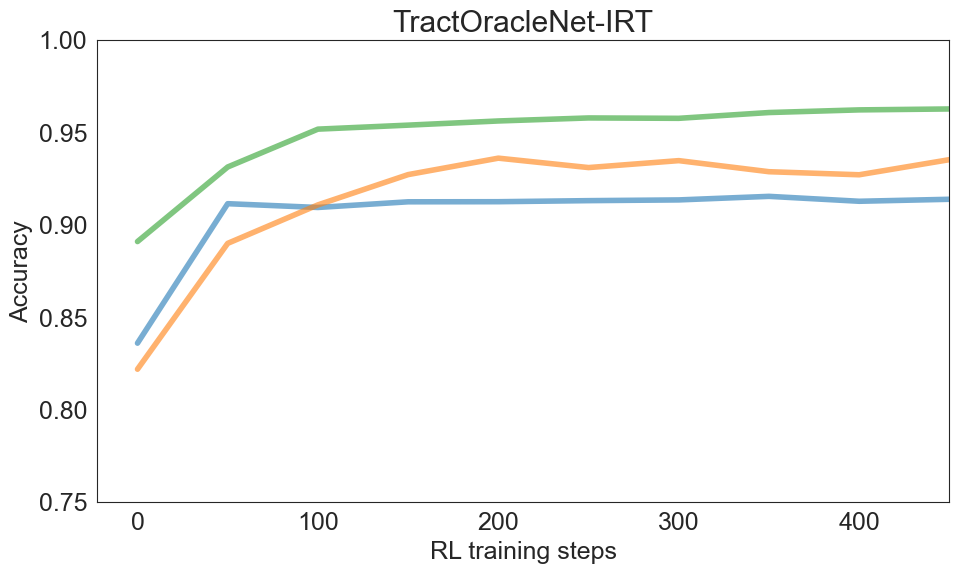}
    \label{fig:rlhf-acc-results}

    \caption{Accuracy of the reward network during the training steps without IRT (left) and with IRT (right) with RecobundlesX as a reference.}
    \label{fig:rlhf-acc-performance}
\end{figure}

\subsubsection{In vivo results : HCP and TractoInferno}

We report in Table~\ref{tab:tractoinferno-hcp-all} the number of streamlines deemed anatomically plausible by each filtering algorithm for each experiment. In each experiment, in order to emphasize which reference filtering method was used in the agents' training, we highlight the corresponding column in gray. We quickly notice, that all proposed methods significantly outperform the four baseline methods, producing on average between 3$\times$ and 20$\times$ more valid streamlines than sd\_stream, ifod2 and ptt, across both datasets. We also see how the use of an oracle drastically improves the performance of baseline RL methods such as Track-to-Learn, in almost all scenarios. Now, we notice that, as the length of the training extends, the agent specializes on tracking according to the inherent anatomical constraints posed by the different reference methods. A good example of this would be using the extractor\_flow as a reference for TractoInferno (i.e. \textit{EXT-ref}) where the number of plausible streamlines is high both for RecobundlesX and extractor\_flow when tracking using \textit{SAC-3K}. However, as the performance keeps increasing for extractor\_flow when introducing the IRT scheme, the number of plausible streamlines for RecobundlesX crumbles to less than half the number from \textit{SAC-3K}. Other similar scenarios in our experiments from table ~\ref{tab:tractoinferno-hcp-all}, but in most cases the number of plausible streamlines increase even when the filtering algorithm isn't used as a reference for the corresponding experiment. We notice that, specifically for the \textit{RBX-ref} experiment on HCP, the number of plausible streamlines remains essentially the same across the different training regimes for the oracle-based RL agents, but still exceeds the other non-oracle methods by a factor of up to 4.

\begin{figure}
    \centering
    \includegraphics[width=1.0\linewidth]{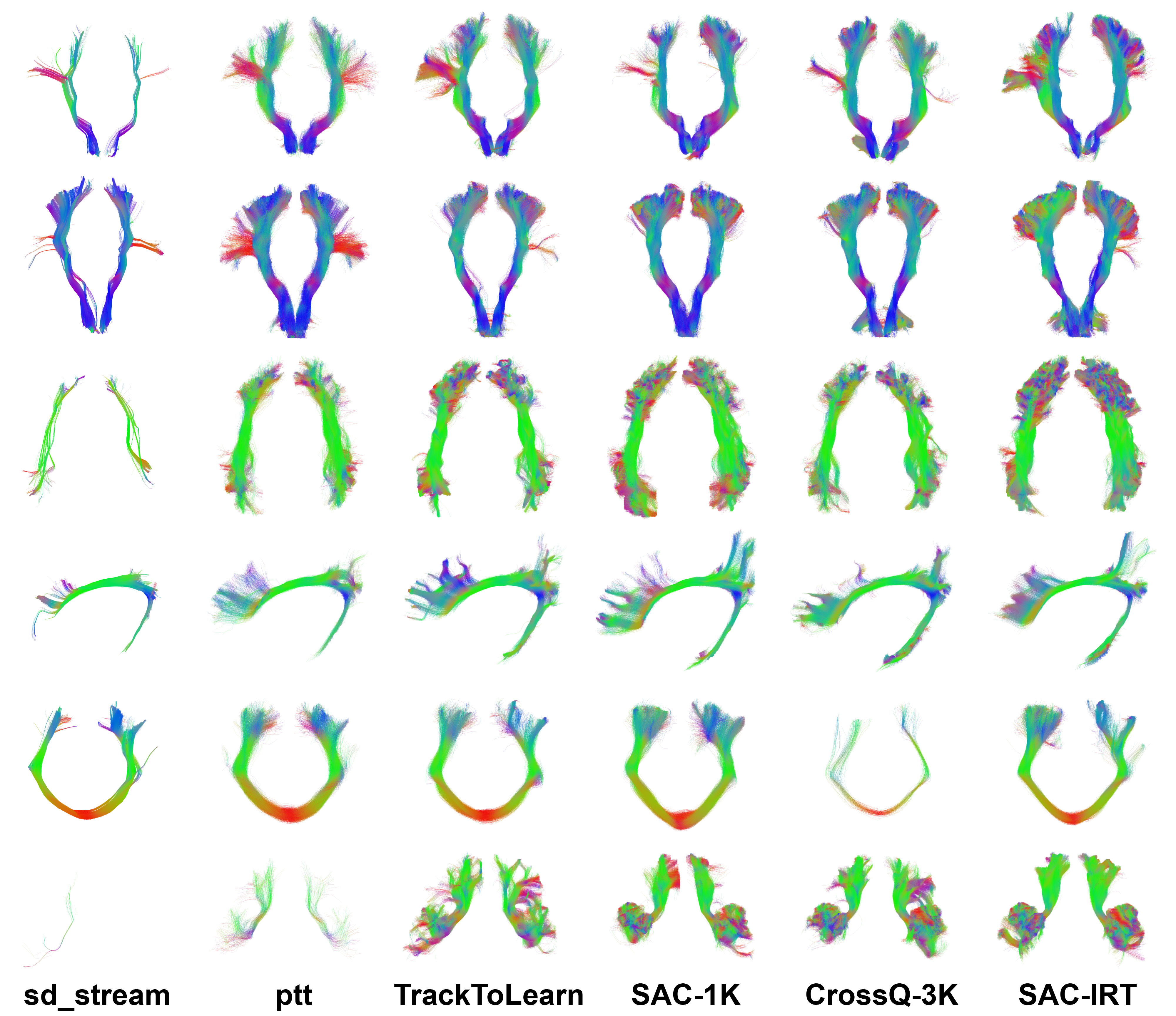}
    \caption{Visualization of the parieto-occipito pontine tract (1st row), pyramidal tract (2nd row), inferior longitudinal fasciculus (3rd row), left cingulum (4th row), occipital lobe of the corpus callosum (5th row), right and left uncinate fasciculus (last row) for the \textbf{subject 1006 of the TractoInferno dataset}. For all methods, results were produced by tracking with 5M seeds each (distributed uniformly across the seeding mask).}
    \label{fig:results_irt_tracto}
\end{figure}

Figure \ref{fig:results_irt_tracto} displays a few bundles from subject 1006 of the TractoInferno dataset which were tracked using baseline methods as well as the proposed methods which were trained on the same dataset. For the sake of conciseness, we selected a subset of the different methods which also reflects the performance of the other variants. With this figure, we showcase the qualitative improvements of our proposed methods compared to the baseline methods. Visually, our proposed methods produce more anatomically plausible and densely reconstructed bundles with clear fanning and improved spatial coherence. This is particularly evident for the IRT-based method, which, as reported in Tables~\ref{tab:tractoinferno-hcp-all} and ~\ref{tab:transfer-learning}, generally yields the highest number of true positive streamlines among all tested approaches. We also notice that some bundles (e.g. the unicinate fasciculus) are more easily reconstructed by RL-based methods and more clearly structured when using the IRT training procedure.

\begin{table}
    \centering
    \tiny
    \caption{Total number of streamlines recovered by RecobundlesX (RBX)~\cite{rheault2020analyse}, extractor\_flow (EXT)~\cite{Petit23} and Verifyber (VER)~\cite{astolfi2023supervised} on the TractoInferno and HCP datasets for all tracking algorithms. Proposed RL agents were rewarded with different oracles across 3 experiments. Each experiment denoted by the headers RBX-ref, EXT-ref and VER-ref were conducted using the same procedures with the exception of the reference method used to score the streamlines during the oracles' initial training was RecobundlesX, extractor\_flow and Verifyber respectively. The grayed columns highlights the reference method used to train the oracle for each experiment. The number of streamlines should be maximized and the best result is highlighted in \textbf{bold}.}
    
    \begin{tabular}{ll | >{\columncolor[gray]{0.9}}ccc | c>{\columncolor[gray]{0.9}}cc | cc>{\columncolor[gray]{0.9}}c}
        \hline
         &  & \multicolumn{3}{c}{\textit{RBX-ref}} & \multicolumn{3}{c}{\textit{EXT-ref}} & \multicolumn{3}{c}{\textit{VER-ref}} \\
         & \textbf{Method} & \textbf{RBX} & \textbf{EXT} & \textbf{VER} & \textbf{RBX} & \textbf{EXT} & \textbf{VER} & \textbf{RBX} & \textbf{EXT} & \textbf{VER} \\
         \hline
         \multirow{9}{*}{\rotatebox{90}{\textbf{TractoInferno}}}
         & sd\_stream & 1,6M & 6,0M & 56,6M & 1,6M & 6,0M & 56,6M & 1,6M & 6,0M & 56,6M \\
         & ifod2 & 8,0M & 9,7M & 83,0M & 8,0M & 9,7M & 83,0M & 8,0M & 9,7M & 83,0M \\
         & ptt & 8,7M & 16,0M & 82,7M & 8,7M & 16,0M & 82,7M & 8,7M & 16,0M & 82,7M \\
         & TTL & 9,2M & 12,4M & 84,3M & 9,2M & 12,4M & 84,3M & 9,2M & 12,4M & 84,3M \\ \cline{2-11}
         & SAC-1K & 20,8M & \textbf{21,0M} & 103,7M & 10,4M & 22,2M & 110,6M & 8,0M & 19,2M & 117,6M \\
         & SAC-3K & 21,6M & 12,8M & 93,0M & \textbf{29,8M} & 24,4M & 110,2M & 17,4M & 19,4M & 118,5M \\
         & CrossQ-3K & 21,4M & 11,9M & 81,3M & 25,6M & 23,5M & 110,3M & \textbf{21,5M} & \textbf{21,6M} & 122,6M \\
         & SAC-IRT & \textbf{33,4M} & 15,3M & \textbf{104,7M} & 13,6M & 24,9M & \textbf{116,7M} & 16,9M & 16,2M & 120,4M \\
         & CrossQ-IRT & 31,7M & 15,0M & 95,8M & 15,4M & \textbf{25,2M} & 111,7M & 16,7M & 18,6M & \textbf{123,8M} \\
         \hline
         \hline
         \multirow{9}{*}{\rotatebox{90}{\textbf{HCP}}}
         & sd\_stream & 688K & 880K & 5,8M & 687,5K & 879,9K & 5,8M & 687,5K & 880K & 5,8M \\
         & ifod2 & 769,7K & 1,5M & 12,0M & 769,7K & 1,5M & 12,0M & 769,7K & 1,5M & 12,0M \\
         & ptt & 2,1M & 2,1M & 11,4M & 2,1M & 2,1M & 11,4M & 2,1M & 2,1M & 11,4M \\
         & TTL & 2,2M & 2,3M & 11,4M & 2,2M & 2,3M & 11,3M & 2,2M & 2,3M & 11,3M \\ \cline{2-11}
         & SAC-1K & 7,7M & 2,8M & 12,9M & 1,9M & 2,9M & 13,9M & 1,3M & 2,3M & 14,5M \\
         & SAC-3K & 7,0M & 2,2M & \textbf{13,5M} & 3,1M & 3,6M & 13,5M & 4,3M & 2,5M & 15,4M \\
         & CrossQ-3K & \textbf{8,4M} & \textbf{2,9M} & 13,4M & 3,3M & 3,4M & 14,3M & 3,7M & 2,9M & 15,3M \\
         & SAC-IRT & 7,4M & 2,7M & 12,5M & 4,5M & \textbf{4,0M} & 14,2M & \textbf{7,5M} & 2,9M & \textbf{16,4M} \\
         & CrossQ-IRT & 7,4M & 2,6M & 12,8M & \textbf{4,6M} & 3,9M & \textbf{14,6M} & 6,5M & \textbf{3,2M} & 16,0M
    \end{tabular}
    \label{tab:tractoinferno-hcp-all}
\end{table}

\subsubsection{Transfer learning: Penthera-3T, TractoInferno and HCP}

Table~\ref{tab:transfer-learning} presents the number of streamlines recovered by RecoBundlesX in a transfer learning setting. Agents trained on the HCP dataset were evaluated on the TractoInferno datasets, while agents trained on TractoInferno were tested on both the HCP and Penthera-3T datasets.

\begin{figure}
    \centering
    \includegraphics[width=1.0\linewidth]{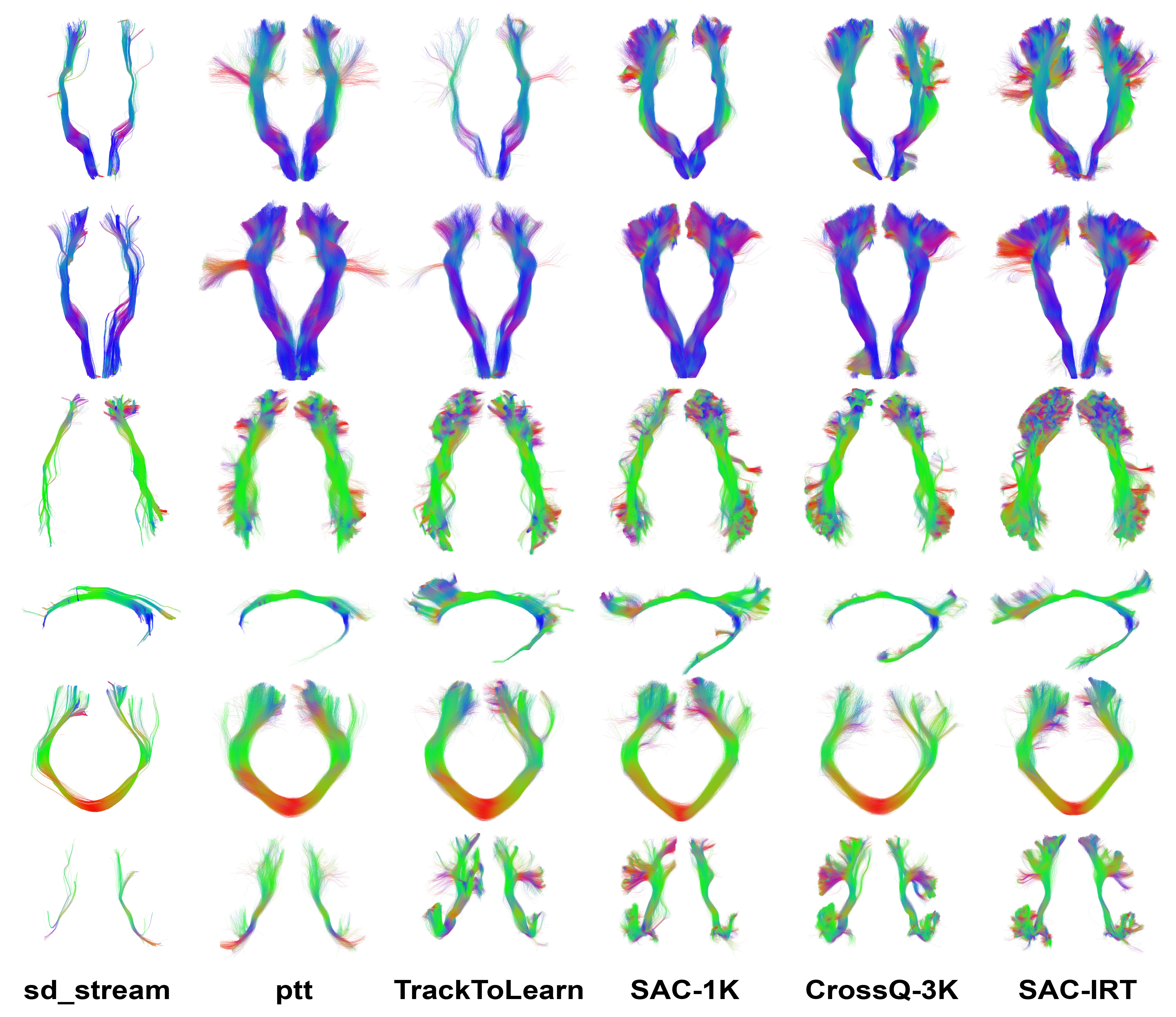}
    \caption{Visualization of the transfer learning experiment by \textbf{tracking on the Penthera-3T dataset after training on TractoInferno}. Visualizations are of the parieto-occipito pontine tract (1st row), pyramidal tract (2nd row), inferior longitudinal fasciculus (3rd row), left cingulum (4th row), occipital lobe of the corpus callosum (5th row), right and left uncinate fasciculus (last row) on the first scan of the first session of the subject 01 from the Penthera-3T dataset. For all methods, results were produced by tracking with 4,5M seeds each (distributed uniformly across the seeding mask).}
    \label{fig:results_irt_penthera}
\end{figure}

We observe that the agents trained on TractoInferno generalize very well on both the HCP and Penthera-3T datasets as all oracle-based methods produce an average between 2 to 7 times more plausible streamlines than baseline methods. Those generalization results are illustrated in figure \ref{fig:results_irt_penthera}. We notice that for \textit{SAC-IRT}, especially for the parieto-occipito pontine tracts, the pyramidal tracts and the inferior longitudinal fasciculus, the bundles are much more dense and have wider fanning than other methods. For other illustrated bundles, SAC-IRT and other oracle-based algorithms are on-par or better than other non oracle-based algorithms in terms of bundle density, fanning and spatial coherence.

Although the generalization capabilities of the \textit{SAC-IRT} and \textit{CrossQ-IRT} algorithms trained on the HCP dataset do not standout as much as in Table \ref{tab:transfer-learning} compared to other oracle-based methods, they are on-par with them. Overall, all the oracle-based algorithms still outperform the other tractography methods.

\begin{table}
    \centering
    \footnotesize
    \caption{Total number of streamlines recovered by RecobundlesX~\cite{rheault2020analyse} in a \textbf{transfer learning} setting for the TractoInferno, HCP, and Penthera-3T datasets using all previous agents. \textbf{HCP $\rightarrow$ TractoInferno} indicate that the agent was trained on HCP and tested on TractoInferno. Similarily, for the entry \textbf{TractoInferno $\rightarrow$ Penthera-3T}, the agent was trained on TractoInferno while the tracking was done on the Penthera-3T dataset and so on. Results from sd\_stream, ifod2 and ptt are reported from table \ref{tab:tractoinferno-hcp-all}. Results in \textbf{bold} outperform others.}

    \begin{tabular}{lccc}
        \toprule
         \textbf{Method} & \shortstack{\textbf{HCP}\\$\downarrow$\\ \textbf{TractoInferno}} & \shortstack{\textbf{TractoInferno} \\ $\downarrow$ \\ \textbf{HCP}} & \shortstack{\textbf{TractoInferno} \\ $\downarrow$ \\ \textbf{Penthera-3T}} \\
        \midrule
        sd\_stream     & 1,6M & 687,5K & 3,6M \\
        ifod2          & 8,0M & 769,7K & 3,4M \\
        ptt            & 8,7M & 2,1M & 8,7M \\
        TTL   & 7,7M & 1,7M & 8,1M \\
        \midrule
        SAC-1k         & 5,4M & 3,9M & 14,0M \\
        SAC-3k         & \textbf{11,0M} & 3,0M & 14,0M \\
        CrossQ-3k      & 4,7M & 3,5M & 14,0M \\
        SAC-IRT        & 8,9M & \textbf{5,7M} & \textbf{22,2M} \\ 
        CrossQ-IRT     & 10,2M & 4,6M & 20,7M \\
        \bottomrule
    \end{tabular}
    \label{tab:transfer-learning}
\end{table}

\section{Discussion}

The results presented in this work highlight two key findings. First, incorporating an oracle into an RL-based tractography system consistently provides a significant advantage. Regardless of the underlying RL algorithm—whether SAC, DroQ, or CrossQ—or whether the system is trained with or without Iterative Reward Training (IRT) or extended spatial context, oracle-guided RL methods systematically outperform both oracle-free RL approaches (such as Track-to-Learn) and traditional tractography techniques like sd\_stream, ifod2 and ptt. The oracle contributes to a more balanced and anatomically accurate streamline generation across evaluation metrics and datasets.

Second, while the ISMRM2015 dataset does not reveal a clear winner among the RL variants, results on the \textit{in vivo} TractoInferno, HCP and Penthera-3T datasets clearly demonstrate the superiority of the IRT scheme. Across the three datasets and all three filtering pipelines, IRT-enhanced methods produce up to 30$\times$ more true positive streamlines than traditional baselines and up to 1.5$\times$ more than their non-IRT RL counterparts.

Additional experimental findings show that extending training to 3,000 episodes (compared to 1,000) does not introduce instability or evidence of reward hacking, further highlighting the robustness of the oracle framework. Furthermore, using shorter streamlines with 32 points—as opposed to 64 or 128—does not compromise the oracle’s accuracy, while substantially reducing computational cost during both training and inference.

The effectiveness of the IRT scheme is further illustrated in Figure~\ref{fig:rlhf-acc-performance}, which tracks the accuracy of the oracle network during RL training on three datasets using RecobundlesX as a reference. Each data point represents the classification performance on streamlines generated by the RL agent every 50 training episodes. The left-hand plots depict training without IRT, while the right-hand plots show results with IRT. In all cases, the reward network was pre-trained on a dataset different from the one used for the agent's training, resulting in initially low accuracy due to domain shift. Nevertheless, the IRT mechanism successfully mitigates this domain gap throughout training, steadily improving the reward model's accuracy. 

We observe from our results in table \ref{tab:ismrm} that the overlap of the methods tend to decrease as we extend the agents' training and as we further increase the number of VC streamlines. We hypothesize that the agents might learn to avoid the edges of the bundles as those edges tend to yield a lower expected cumulative reward in the long run. As there's some stochasticity in the decision making process of an agent, the agents can easily take a wrong turn leading them to exit the tracking mask (potentially in an undesired region), thus prematurely exiting the tracking procedure which potentially lowers the expected reward in the previously visited states. We suspect that tackling this open question could help to propel RL-based tractography algorithms to having a state of the art coverage while having a best in class generation of plausible streamlines.

Building on the insights of this study, several avenues for future research emerge. We here trained several oracles-one for each filtering algorithm-in order to get a fast-performing and relatively light-weight neural network that reproduces the behaviour of only a single filtering method. As each filtering method currently has their own different biases on the anatomy, we could consider having a neural network that is trained to mimic the behaviour of an aggregation of filtering algorithms. Potentially using those filtering methods in a voting system to attribute target scores to a much larger dataset of streamlines. Naturally, to capture the many complex constraints given by the filtering algorithms, a bigger neural network might be needed. As this work didn't explore different oracle architectures, we suspect that the avenue of scaling up the capabilities of the oracle could be an interesting topic to address, especially in a scenario with multiple filtering algorithms. As we limited our approach on scaling the number of points used to represent each streamline to a fixed number of 32, one could probably even resample the streamlines to half that number, making room in the prediction time to scale up the size of the model. Such scaling could also potentially directly benefit our methodology as we found it was harder for the oracle to learn the complex constraints posed by extractor\_flow. Indeed, the oracle is only inputted the geometry (i.e. the directions) of the streamline while the extractor\_flow considered more than only the geometry, causing the learning of the oracle to be harder will less information.

As our results suggest a performance plateau for locally informed tracking strategies, future efforts should focus on developing RL policies capable of integrating more global anatomical context. This could involve incorporating long-range dependencies, using hierarchical policies, or leveraging multi-scale representations of the white matter structure. Additionally, extending the current framework to handle uncertainty and ambiguity in more anatomically complex or degenerated regions remains an open challenge. Another crucial direction involves the systematic validation of these models in pathological cases, such as multiple sclerosis or brain tumors, where structural alterations pose significant challenges to existing tractography methods. Finally, improving the interpretability and clinical usability of RL-based tractography systems will be essential for broader adoption in neuroimaging and neurosurgical planning.

\section{Conclusion}
In this work, we extended the TractOracle-RL framework by incorporating recent advances in reinforcement learning, including CrossQ, DroQ, and an iterative training scheme for the reward network. These enhanced variants were systematically evaluated on a range of diffusion MRI datasets, including both \textit{in silico} and \textit{in vivo} data. Our findings reveal that incorporating anatomical priors via reward-based mechanisms consistently improves tractography performance compared to both traditional methods and oracle-free RL baselines. We found this to be also true when extending the training over a larger number of episodes. This highlights the inherent stability and robustness of oracle-guided approaches. The proposed \textit{Iterative Reward Training} (IRT) scheme not only enhances the oracle's accuracy during training, but also leads to a substantial increase in the number of true positive streamlines generated by the tracking policy, highlighting the importance of incorporating an accurate and robust anatomical signal.

Collectively, these contributions advance the state of RL-based tractography and lay the groundwork for future developments in learning-based white matter reconstruction—particularly in scenarios demanding robust domain adaptation and efficient inference.

\section{Acknowledgement}

This work is in part supported by the Natural Sciences and Engineering Research Council of Canada (RGPIN-2023-04584 and RGPIN-2020-04818) and the institutional USherbrooke Research Chair in Neuroinformatics. 

\footnotesize
\bibliographystyle{elsarticle-num}
\bibliography{references}

\include{elsearticle-appendix}

\end{document}

%% file: elsearticle-appendix.tex
\appendix

\section{Details of reinforcement learning algorithms}\label{app:rl-details}

\subsection{Soft Actor-Critic}\label{sec:sac-appendix}

\begin{algorithm}
\caption{Soft Actor-Critic}\label{alg:sac}
\begin{algorithmic}[1]
\Require{$\theta, \phi_1, \phi_2, \phi'_1, \phi'_2, \alpha, \eta, \bar{\mathcal{H}}$}
\For{every epoch}
    \While {$t \neq T$}
        \State $a_t \sim \pi_\theta(s_t) $
        \State $s_{t+1} \sim p(s_{t+1} | s_{t}, a_t)$ 
        \State $D \gets D \cup (s_t, a_t, r_t, s_{t+1})$
        \State $\{(s_k, a_k, r_k, s_{k+1}\} \gets B$  \hspace{1cm} // B: Replay buffer

        \State $\alpha \gets \alpha \eta \nabla_\alpha J_\pi(\alpha)$
        \State $\theta \gets \theta \eta \nabla_\theta J_\pi(\theta)$
        \State $\phi_1 \gets \eta \nabla \phi_1 J_Q(\phi_1)$
        \State $\phi_2 \gets \eta \nabla \phi_2 J_Q(\phi_2)$
        \State $\phi_1' \gets \tau\phi_1 + (1 - \tau)\phi_1'$
        \State $\phi_2' \gets \tau\phi_2 + (1 - \tau)\phi_2'$   
    \EndWhile
\EndFor
\end{algorithmic}
\end{algorithm}

The Soft Actor-Critic~(SAC) algorithm builds upon the Deterministic Policy Gradient theorem~\cite{silver2014deterministic} which formulates the RL objective as

\begin{equation}
    J_\pi(\theta) = \int_S \rho^\pi Q^{\pi_\theta}_\phi(s, a)ds |_{a=\pi_\theta(s))}
\end{equation}

with $\rho^\pi$ the state-visitation probability density of $\pi$ and $Q^{\pi_\theta}_\phi(\mathbf{s}_t, \mathbf{a}_t) = \mathbb{E}_{\pi_\theta} [r_t + G_{t+1} \mid \mathbf{s}_t, \mathbf{a}_t].$ Therefore, it follows that the policy update can rely solely on $Q$:

\begin{equation}
    \nabla_\theta J_\pi(\theta) = \mathbb{E}_{s\sim\rho^\pi} [ \nabla_aQ^{\pi_\theta}_\phi(s,a)\nabla_\theta \pi_\theta(s) | _{a=\pi_\theta(s)}]
\end{equation}

The Q-function, conversely, is trained to minimize the Temporal Difference~(TD) via mean-squared error:

\begin{align}
    J_Q(\phi) &= \mathbb{E}_{s_t,a_t \sim B}\frac{1}{2}[Q^{\pi_\theta}_\phi(s_t, a_t) - \hat{y}]^2 \\
    \hat{y}_t &= r_t + \gamma Q^{\pi_\theta}_{\phi'}(s_{t+1}, a_{t+1}) \\
    \nabla_\phi J_Q(\phi)&= [Q^{\pi_\theta}_\phi(s_t, a_t) - \hat{y}] \nabla_\phi,
\end{align}
with $\phi'$ the weights of a \emph{target} Q-function used to stabilize training and $\hat{y}$ is the TD target. However, Q-functions tend to suffer from an overestimation bias~\cite{hasselt2010double}. SAC therefore uses two Q-functions and two targets $\phi_1, \phi_2, \phi'_1, \phi'_2$: 

\begin{align}
    \hat{y} &= r_t + \gamma \min(Q^{\pi_\theta}_{\phi_1'}(s_{t+1}, a_{t+1}), Q^{\pi_\theta}_{\phi_2'}(s_{t+1}, a_{t+1}))\\
    \nabla_{\phi_1} J_Q(\phi_1)&= [Q^{\pi_\theta}_{\phi_1}(s_t, a_t) - \hat{y}] \nabla_{\phi_1} \\
    \nabla_{\phi_2} J_Q(\phi_1)&= [Q^{\pi_\theta}_{\phi_2}(s_t, a_t) - \hat{y}] \nabla_{\phi_2}.
\end{align}

Moreover, to fight the usual sensitivity of RL algorithms to their hyperparameters, the authors exploit the Maximum Entropy Reinforcement Learning framework~\cite{haarnoja2017reinforcement} to augment the previously mentioned RL objective with an entropy maximization term:

\begin{equation}
G_t = \sum_{k=t}^{T} \gamma^k \left[ r(\mathbf{s}_{t+k}, \mathbf{a}_{t+k}) + \alpha \mathcal{H}\left( \pi_\theta(\cdot \mid \mathbf{s}_{t+k}) \right) \right].
\end{equation}

Following the updated objective, the Q-function target can be rewritten as:

\begin{align}
    \hat{y} &= r_t + \gamma \min(Q^{\pi_\theta}_{\phi_1'}(s_{t+1}, a_{t+1}), Q^{\pi_\theta}_{\phi_2'}(s_{t+1}, a_{t+1})) - \alpha \log \pi_\theta(s_t | a_t). \\
\end{align}
Simiarly, the policy update becomes:
\begin{equation}
    \nabla_\theta J_\pi(\theta) = \mathbb{E}_{s\sim\rho^\pi} [\nabla_\theta \log \pi_\theta(a | s) + (\nabla_a \log \pi_\theta (a | s) -\nabla_aQ^{\pi_\theta}_\phi(s,a))\nabla_\theta \pi_\theta(s) | _{a=\pi_\theta(s)}]
\end{equation}

Finally, in a follow-up paper~\cite{haarnoja2018bsoft}, the authors demonstrated that the $\alpha$ parameter can be automatically tuned using the following objective:
\begin{equation}
    J(\alpha) = \mathbb{E}_{a_t \sim \pi_{\theta}(s_t)}[ -\alpha \log \pi_\theta(a_t | s_t) - \alpha \bar{\mathcal{H}}],
\end{equation}
with $\bar{\mathcal{H}}$ the desired final entropy. All in all, the SAC algorithm can be summarized by Algorithm~\ref{alg:sac}.

\subsection{DroQ}\label{sec:droq-appendix}

To improve on sample efficiency, the authors of Dropout Q-functions~\cite{hiraoka2021dropout} therefore propose to raise the Update-to-Data~(UTD) ratio of SAC above its standard value of 1. The UTD controls the ratio of optimization steps (i.e. lines 5-12 in algorithm~\ref{alg:sac}) to sample gathering steps (lines 3-4). Sample gathering steps may be computationally expensive in complex scenarios and intuitively, performing more optimization steps should lead to a faster convergence. However, it also empirically leads to a higher overestimation bias by the Q-function(s). Some methods~\cite{hasselt2010double, chen2021randomized} have proposed to use more ($\geq 2$) critics to fight this overestimation but this also leads to more memory and computation needed for each optimization step.

Dropout Q-functions~\cite{hiraoka2021dropout} has been proposed as a way to increase the UTD without amplifying the overestimation bias while keeping a low (2) number of critics.  The authors argue a high number of critics injects model uncertainty into the Q target, which fights overestimation. Instead, the authors include dropout~\cite{srivastava2014dropout} and layer normalization~\cite{ba2016layer} layers in the critics.

\subsection{CrossQ}\label{sec:crossq-appendix}

While a higher UTD does lead to better sample efficiency, it also leads to more computation as more optimization steps are performed. In an effort to improve sample efficiency while reducing computational costs, CrossQ~\cite{bhatt24} was proposed. The algorithm removes target Q functions, which require additional computation and instead uses Batch Normalization~\cite{ioffe2015batch} in the critics' networks.

While Batch Norm. has been tried for RL before, it was either not included in the critic networks~\cite{lillicrap2015continuous, ota2021training} or deemed harmful~\cite{bhatt24}. The authors of CrossQ argue BatchNorm can be useful as long as the distribution of input states and action is respected. Indeed, target networks $Q_{\phi'}$ are often updated via polyak averaging of the actual Q networks $Q_\phi$. However, Q networks are used to compute $Q_{\phi'}(s_{t}, a_{t})$ from "old" policies (stored as transitions in the replay buffer) while target networks are used to compute $Q_{\phi'}(s_{t+1}, a_{t+1})$, with $a_{t+1} \sim \pi_\theta(s_{t+1})$ the current policy. The authors argue the running statistics for $s_t, a_t$ and $s_{t+1}, a_{t+1}$ do not match and degrade the quality of predictions from networks with BatchNorm layers.

Instead, the authors propose to match the statistics of $s_t, a_t$ and $s_{t+1}, a_{t+1}$ by feeding them jointly to the same network: target Q networks are omitted and $s_t, s_{t+1}$, $a_t, a_{t+1}$ are concatenated batch-wise and input to the Q functions. This trick ensures that Batch Norm's normalizing moments occur from the union of both batches, corresponding to an equal contribution of both batches in calculating the normalization running statistics. 

The Q-function objective and update can therefore be re-written as
\begin{align}
    [\hat{q_t}, \hat{q}_{t+1}] &= Q^{\pi_\theta}_\phi([s_{t}, s_{t+1}], [a_t, a_{t+1}]) \\ 
    \hat{y} &= r_t + \gamma \hat{q}_{t+1}\\
    \nabla_{\phi} J_Q(\phi)&= [\hat{q_t} - \hat{y}] \nabla_{\phi}.
\end{align}

\section{Oracle architecture}\label{sec:oracle-architecture}

\begin{figure}[h!]
    \centering
    \includegraphics[width=0.5\linewidth]{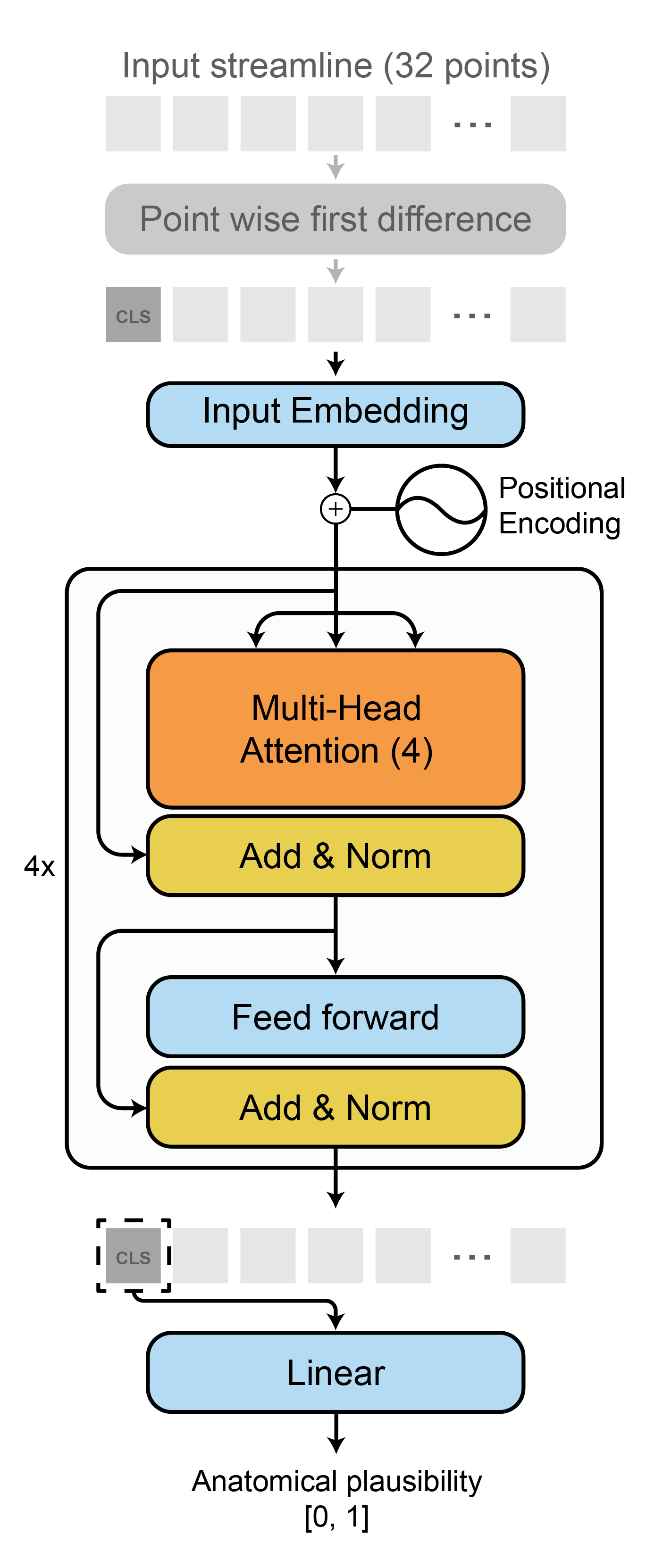}
    \caption{Architecture of the oracle used in our experiments that is directly inspired by \cite{vaswani2017attention}. Inputs in this figure are streamlines represented by 32 (x, y, z) points. For more stable and accurate predictions, the oracle predicts on the directions between each point of the streamline rather than predicting on the (x, y, z) values of the points. This transformer is composed of 4 transformer blocks each having 4 multi-head attention blocks.}
    \label{fig:oracle-architecture}
\end{figure}

Figure~\ref{fig:oracle-architecture} offers an overview of the oracle's architecture. The streamlines resampled to a fixed number of points (e.g. 32, 64 or 128 points) are initially transformed into 31 directions by performing a first difference along each neighbouring point. This trips off any position information to only keep the general geometry of the streamline that is used to predict its anatomical plausibility. Each direction (and the CLS token) is then embedded into a vector of size 32 that will then be positionally encoded before being fed to the 4 transformer encoding blocks. From the output of the transformer, we use the CLS token as input to the prediction head (a linear layer of size 32) followed by a sigmoid which produces our prediction value ranging from 0 to 1. For experiments of section \ref{sec:oracle-nb-points}, the architecture stays the same, but the input would change to 64 or 128 points.

\section{fODF neighbourhood autoencoder}\label{sec:fodf-ae-impl}

\begin{figure}
    \centering
    \includegraphics[width=1.0\linewidth]{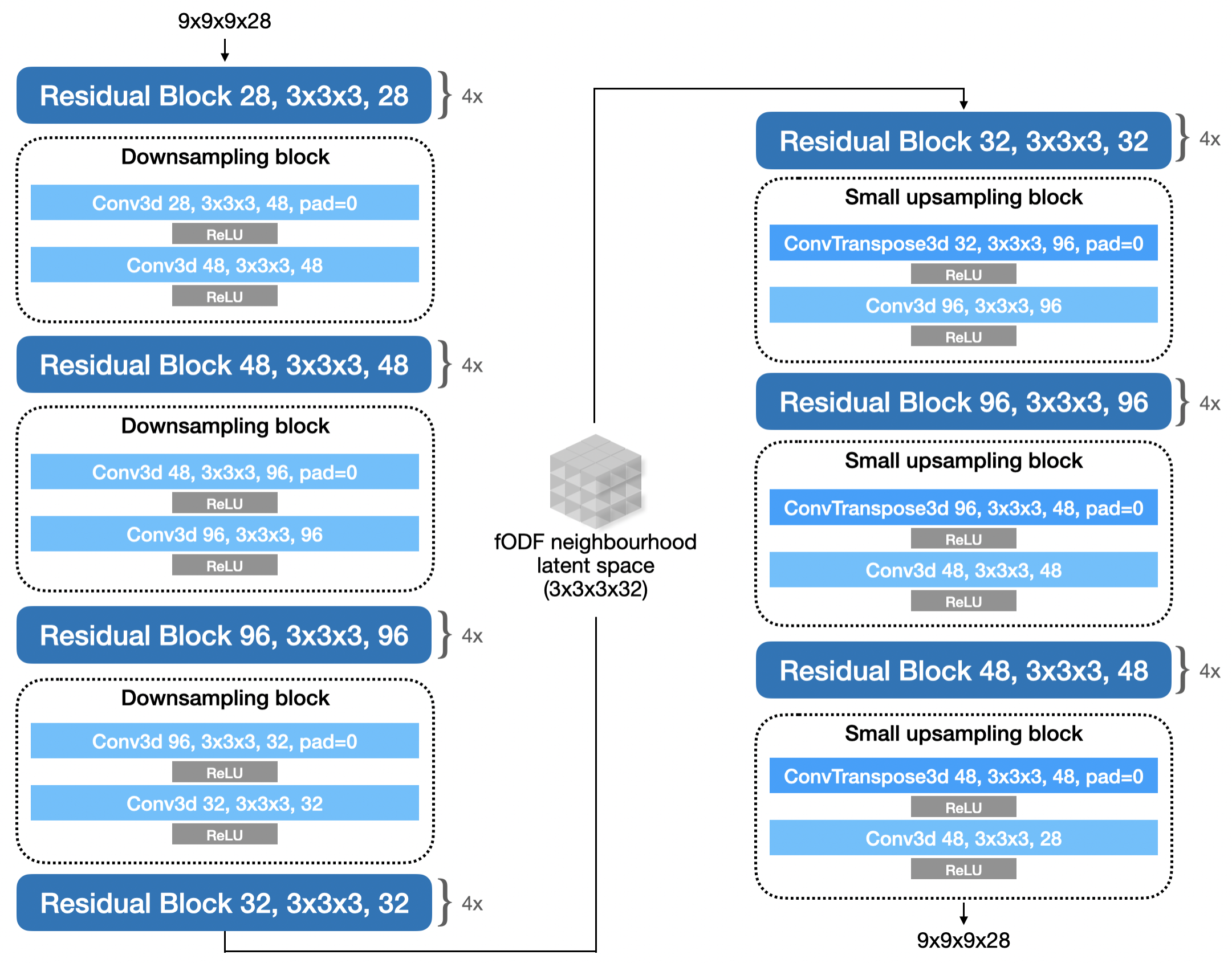}
    \caption{Architecture of the fODF neighbourhood autoencoder. This network takes a patch of 9x9x9x28 from the ISMRM2015 diffusion signal volume as input, compresses it to a latent representation of shape 3x3x3x32 and learned to faithfully reconstruct the input. Each block holding a convolution is suffixed with "input channels, 3d filter size, output channels". The residual block is inspired from \cite{he2016deep} and is simply a Conv3D-BatchNorm-ReLU-Conv3D-BatchNorm block where the number of channels is constant and the input is summed to the output of those layers.}
    \label{fig:fodf-ae-architecture}
\end{figure}

The encoder used to encode a wider diffusion signal neighbourhood is a convolutional autoencoder, as shown in figure \ref{fig:fodf-ae-architecture}, which was trained to compress and reconstruct patches of 9x9x9 voxels from the diffusion signal. As it is only used as an experiment for the ISMRM2015 dataset, it was trained on patches randomly selected between all the possible patches that have the central voxel within the white matter region. By learning in a self-supervised way to compress and reconstruct different patches, the encoder eventually learns to extract meaningful representations of the input into a latent space that is approximately 23x smaller than the input. Once the self-supervised training is done, we leverage and meaningful representations learned by the encoder as new inputs for the RL agents so that each state encapsulate a wider context that can be leveraged by the agent.

We chose to decouple the representation learning from the RL training loop as it has been shown that this strategy helped learning in complex scenarios \cite{ota2020can, ota2021training}. In addition, since the encoder parameters are fixed, this allows to reduce considerably the number of parameters to optimize during the training loop which consequentially also limits the time and the computational resources needed to run the RL training loop.

\section{List of acronyms for ISMRM2015 metrics}\label{sec:tractometer-metrics}

\begin{samepage}
\begin{itemize}
    \item VC: percentage of streamlines exhibiting a valid connection.
    \item VB: number of valid bundles (max 21).
    \item IC: percentage of streamlines exhibiting an invalid connection.
    \item IB: number of invalid bundles connecting regions that should not be connected.
    \item OL: Overlap (percentage of ground truth voxels recovered).
    \item OR: Overreach. Quantifies how the algorithm extends past the volume of the ground-truth (percentage of false positive voxels).
    \item F1: F1-Score, equivalent to the Dice score. The higher the value (max 1), the more the reconstruction volume matches the volume of the ground truth.
    \item NC: percentage of streamlines that do not connect two gray-matter regions.
\end{itemize}
\end{samepage}

\section{Experiments hyperparameters}
\begin{table}[h!]
    \centering
    \scriptsize
    \caption{\textbf{TractOracle-Net training configuration}. Key hyperparameters which were used to initially train the oracles. The same hyperparameters were used to train oracles on the different references (i.e. Tractometer, Recobundles, extractor\_flow or Verifyber). Only the data and the labels change between experiments.}
    \begin{tabular}{ll}
        \hline
        \textbf{Hyperparameter} & \textbf{Value} \\
        \hline
        Oracle Train Steps (nb epochs)  & 50 \\
        Oracle Learning Rate            & 0.0005 \\
        Oracle Batch Size               & 1024 \\
        \hline
    \end{tabular}
    \label{tab:hyperparams-oraclenet}
\end{table}

\begin{table}[h!]
    \centering
    \scriptsize
    \caption{\textbf{RL training configuration, ISMRM2015}. Key hyperparameters which were used in all RL-based experiments. When performing iterative reward tuning, the extra configuration is used with the rest of the configuration that's normally used.}
    \begin{tabular}{ll}
        \hline
        \textbf{Hyperparameter} & \textbf{Value} \\
        \hline
        \multicolumn{2}{c}{\textit{Experiment parameters}} \\
        \hline
        Number of seed per voxel (NPV) & 2 \\
        Min. Streamline Length      & 20 \\
        Max. Streamline Length      & 200 \\
        Binary Stopping Threshold   & 0.1 \\
        Angle stopping criterion (degrees) & 30 \\
        RNG seed                    & 1111 \\
        \hline
        \multicolumn{2}{c}{\textit{Training configuration (SAC, CrossQ)}} \\
        \hline
        Learning Rate               & 0.0005 \\
        Max Episodes                & (1000 $\vert$ 3000) \\
        Replay Buffer Size          & 1,000,000 \\
        Number of Directions        & 100 (4 for TrackToLearn experiments) \\
        Discount Factor ($\gamma$)  & 0.95 (0.75 for TrackToLearn experiments) \\
        Initial entropy coefficient ($\alpha_{\text{init}}$) & 0.2 \\
        Update-to-Data Ratio (UTD)  & 1 (5 for DroQ) \\
        Hidden Dimensions           & 1024-1024-1024 \\
        Batch Size                  & 4096 \\
        Number of Actors            & 4096 \\
        Oracle Bonus                & 10 (0 for TrackToLearn experiments) \\
        Neighborhood Radius         & 1 (grid of radius 9 for CrossQ-AE) \\
        \hline
        \multicolumn{2}{c}{\textit{Extra configuration for IRT training}} \\
        \hline
        Number of IRT iterations    & 60 \\
        Warmup Agent Steps          & 150 (SAC) $\vert$ 200 (CrossQ) \\
        Agent Train Steps           & 50 \\
        Oracle Train Steps (1st IRT iter.) & 5 \\
        Oracle Train Steps (per IRT iter.) & 1 \\
        Oracle Learning Rate        & 0.0005 \\
        Nb streamlines per iter.    & 250k \\
        Max. dataset size           & 4M \\
        IRT tracking NPV            & 2 \\
        Oracle Batch Size           & 1024 \\
        Reference used for oracle training & Tractometer \\
        \hline
    \end{tabular}
    \label{tab:rl-hyperparams-ismrm}
\end{table}

\begin{table}[h!]
    \centering
    \scriptsize
    \caption{\textbf{RL training configuration, \textit{in-vivo} datasets}. Key hyperparameters which were used in all RL-based experiments. When performing iterative reward tuning, the extra configuration is used with the rest of the configuration that's normally used.}
    \begin{tabular}{ll}
        \hline
        \textbf{Hyperparameter} & \textbf{Value} \\
        \hline
        \multicolumn{2}{c}{\textit{Experiment parameters}} \\
        \hline
        Number of seed per voxel (NPV) & 2 \\
        Min. Streamline Length      & 20 \\
        Max. Streamline Length      & 200 \\
        Binary Stopping Threshold   & 0.1 \\
        Angle stopping criterion (degrees) & 30 \\
        RNG seed                    & 1111 \\
        \hline
        \multicolumn{2}{c}{\textit{Training configuration (SAC, CrossQ)}} \\
        \hline
        Learning Rate               & 0.0005 \\
        Max Episodes                & (1000 $\vert$ 3000) \\
        Replay Buffer Size          & 1,000,000 \\
        Number of Directions        & 100 (4 for TrackToLearn experiments) \\
        Discount Factor ($\gamma$)  & 0.95 (0.75 for TrackToLearn experiments) \\
        Initial entropy coefficient ($\alpha_{\text{init}}$) & 0.2 \\
        Update-to-Data Ratio (UTD)  & 1 \\
        Hidden Dimensions           & 1024-1024-1024 \\
        Batch Size                  & 4096 \\
        Number of Actors            & 4096 \\
        Oracle Bonus                & 10 (0 for TrackToLearn experiments) \\
        Neighborhood Radius         & 1 \\
        \hline
        \multicolumn{2}{c}{\textit{Extra configuration for IRT training}} \\
        \hline
        Number of IRT iterations    & 60 \\
        Warmup Agent Steps          & 150 (SAC) $\vert$ 200 (CrossQ) \\
        Agent Train Steps           & 50 \\
        Oracle Train Steps (1st IRT iter.) & 5 \\
        Oracle Train Steps (per IRT iter.) & 1 \\
        Oracle Learning Rate        & 0.0005 \\
        Nb streamlines per iter.    & 250k \\
        Max. dataset size           & 4M \\
        IRT tracking NPV            & 2 \\
        Oracle Batch Size           & 1024 \\
        Reference used for oracle training & (Recobundles $\vert$ extractor\_flow $\vert$ Verifyber) \\
        \hline
    \end{tabular}
    \label{tab:rl-hyperparams-in-vivo}
\end{table}